\pdfoutput=1
\documentclass[11pt]{article}
\usepackage[T1]{fontenc}
\usepackage{lmodern}
\usepackage[margin=1in]{geometry}
\usepackage{amsmath,amsthm,amssymb}
\usepackage{booktabs}
\usepackage{enumitem}
\usepackage{placeins}
\usepackage{tabularx}
\usepackage[numbers,sort&compress]{natbib}
\usepackage{xcolor}
\usepackage{graphicx}
\usepackage[colorlinks=true,linkcolor=blue!50!black,citecolor=blue!50!black,urlcolor=blue!50!black]{hyperref}
\theoremstyle{plain}
\newtheorem{theorem}{Theorem}
\newtheorem{proposition}{Proposition}
\newtheorem{lemma}{Lemma}
\newtheorem{corollary}{Corollary}[theorem]
\newtheorem{alemma}{Lemma}

\theoremstyle{definition}

\newtheorem{problem}{Problem}
\theoremstyle{remark}
\newtheorem{remark}{Remark}

\newcommand{\R}{\mathbb{R}}
\newcommand{\V}{\mathcal{V}}
\newcommand{\D}{\mathcal{D}}
\newcommand{\F}{\mathcal{F}}
\newcommand{\Hh}{\hat{H}}
\newcommand{\rank}{\operatorname{rank}}
\newcommand{\rk}[1]{\operatorname{rank}_{#1}}
\newcommand{\cut}{(L|R)}
\newcommand{\fro}[1]{\lVert #1 \rVert_F}
\newcommand{\opn}[1]{\lVert #1 \rVert_2}
\newcommand{\Cx}{C^{F}}
\newcommand{\IB}{I_{\mathrm{B}}}
\newcommand{\ISh}{I_{\mathrm{Sh}}}
\newcommand{\one}{\mathbf 1}
\newcommand{\RepoURL}{\url{https://github.com/oguzhankatar/CutRank}}
\newcommand{\NInst}{6{,}720}

\newcommand{\TotalViol}{0}
\newcommand{\RunTime}{18}
\newcommand{\TableOne}{%
dense\_int & 560 & 0 & 0 & 560 & 0 & -- & 560/560\\
sparse\_int & 560 & 0 & 157 & 397 & 6 & 0/560 & 560/560\\
pairwise\_int & 560 & 0 & 0 & 173 & 387 & 0/560 & 560/560\\
separable\_int & 560 & 0 & 0 & 20 & 540 & 0/560 & 560/560\\
lowrank\_int & 560 & 0 & 0 & 176 & 384 & -- & 560/560\\
multiplicative\_int & 560 & 0 & 0 & 560 & 0 & 0/560 & 560/560\\
unanimity & 560 & 0 & 424 & 136 & 0 & 0/560 & 560/560\\
dense\_gauss & 560 & 0 & 0 & 560 & 0 & -- & --\\
decay & 560 & 0 & 0 & 560 & 0 & -- & --\\
lowrank\_noisy & 560 & 0 & 0 & 560 & 0 & -- & --\\
multiplicative\_float & 560 & 0 & 0 & 560 & 0 & 0/560 & --\\
sparse\_float & 560 & 0 & 176 & 364 & 20 & 0/560 & --\\
}
\newcommand{\EpsNonVac}{10{,}790}
\newcommand{\TableTwo}{%
Thm.~\ref{thm:numerical} ($\operatorname{rank}_\tau$, M\"obius) & 86{,}767 & 0 & 0.068 & 0.099\\
Prop.~\ref{prop:epsrank} ($\varepsilon$-rank, M\"obius) & 23{,}584 & 0 & -- & --\\
Thm.~\ref{thm:fourier} ($\operatorname{rank}_\tau$, Fourier) & 96{,}655 & 0 & 0.238 & 0.140\\
}
\newcommand{\SvSandwichViol}{0}
\newcommand{\InterViol}{0}
\newcommand{\ExpMax}{0.928}
\newcommand{\SvErrF}{\le 10^{-13}}
\newcommand{\BzN}{300}
\newcommand{\BzErr}{\le 10^{-14}}
\newcommand{\BzErrDiv}{\le 10^{-14}}
\newcommand{\BzRank}{0}
\newcommand{\FourN}{6{,}720}
\newcommand{\FourViol}{0}
\newcommand{\FourLamViol}{0}

\newcommand{\FourMed}{0.267}
\newcommand{\FourNinety}{0.680}
\newcommand{\FourMax}{1.0000}
\newcommand{\FourGzeroN}{824}
\newcommand{\FourGzeroMax}{0.6931}
\newcommand{\TightSTwo}{0.693147}
\newcommand{\MaxRatioPos}{0.9534}
\newcommand{\TableTight}{%
0.1 & 0.100000 & 0.903470 & 0.903470 & 0.903868\\
0.3 & 0.300000 & 1.400590 & 1.400590 & 1.406497\\
0.5 & 0.500000 & 2.047693 & 2.047693 & 2.079442\\
0.7 & 0.700000 & 2.939291 & 2.939291 & 3.101093\\
0.9 & 0.900000 & 4.013806 & 4.013806 & 5.298317\\
}
\newcommand{\PFiveN}{36}
\newcommand{\PFiveViol}{0}
\newcommand{\TableFive}{%
1 & 1 & 4 & 0.33333 & 0.33333 & 0.25131 & 0.57536 & 2\\
2 & 2 & 6 & 0.60000 & 0.60000 & 0.08338 & 0.12908 & 2\\
3 & 3 & 8 & 0.77778 & 0.77778 & 0.02500 & 0.03150 & 2\\
4 & 4 & 10 & 0.88235 & 0.88235 & 0.00694 & 0.00783 & 2\\
5 & 5 & 12 & 0.93939 & 0.93939 & 0.00184 & 0.00195 & 2\\
6 & 6 & 12 & 0.96923 & 0.96923 & 0.00047 & 0.00049 & 2\\
}
\newcommand{\PFiveNmin}{4}
\newcommand{\PFiveNmax}{12}
\newcommand{\BzSvViol}{0}

\newcommand{\FiveChecks}{40{,}320}
\newcommand{\FiveViol}{0}
\newcommand{\FiveInfChecks}{26{,}260}
\newcommand{\FiveInfViol}{0}
\newcommand{\FiveAttChecks}{84}
\newcommand{\FiveAttViol}{0}
\newcommand{\TableGrowth}{%
2 & 4 & 1.3437 & 1.3040 & 1.2686 & 1.2379 & 1.1903 & 1.0498\\
3 & 8 & 1.8524 & 1.6336 & 1.4629 & 1.3471 & 1.2238 & 1.0498\\
4 & 16 & 2.3466 & 1.8878 & 1.5616 & 1.3806 & 1.2274 & 1.0498\\
5 & 32 & 2.8622 & 2.1164 & 1.6233 & 1.3943 & 1.2280 & 1.0498\\
6 & 64 & 3.4080 & 2.3342 & 1.6648 & 1.4006 & 1.2281 & 1.0498\\
}
\newcommand{\GrowthLimOneHalf}{1.7632}
\newcommand{\GrowthLimTwo}{1.4065}
\newcommand{\GrowthLimThree}{1.2282}
\newcommand{\GrowthLimInf}{1.0498}
\newcommand{\DegCases}{506}
\newcommand{\DegGenViol}{0}
\newcommand{\DegIntViol}{0}
\newcommand{\DegIntEq}{500}
\newcommand{\DegCasesThirty}{8{,}555}
\newcommand{\DegOrderViol}{0}
\newcommand{\DegStrict}{479}
\newcommand{\DegFirstL}{6}

\newcommand{\DegFirstK}{5}
\newcommand{\DegFirstRho}{42}
\newcommand{\DegFirstGrouped}{47}
\newcommand{\DegFirstRow}{56}
\newcommand{\SymN}{300}
\newcommand{\SymViol}{0}

\newcommand{\SymGenMiss}{0}
\newcommand{\ThrN}{588}
\newcommand{\ThrViol}{0}
\newcommand{\DickeN}{588}
\newcommand{\DickeViol}{0}
\newcommand{\GloveN}{36}
\newcommand{\GloveViol}{0}
\newcommand{\MajN}{12}
\newcommand{\MajQ}{7}
\newcommand{\MajNnz}{1{,}586}
\newcommand{\MajTot}{4{,}096}

\newcommand{\MajProfile}{2, 3, 4, 5, 6, 6, 6, 5, 4, 3, 2}
\newcommand{\GloveSixNnz}{3{,}969}
\newcommand{\GloveSixChi}{6}
\newcommand{\AddN}{200}
\newcommand{\AddViol}{0}
\newcommand{\PwN}{300}
\newcommand{\PwViol}{0}
\newcommand{\PwDP}{75}

\newcommand{\PwChainN}{43}
\newcommand{\PwChainViol}{0}
\newcommand{\PwGenN}{300}
\newcommand{\PwGenViol}{0}
\newcommand{\ProfViol}{0}
\newcommand{\SloccErr}{\le 10^{-15}}
\newcommand{\TableProfiles}{%
unanimity $u_T$, $T=\{2,3,5\}$ & 1, 1, 1, 1, 1, 1, 1 & 0, 1, 1, 1, 0, 0, 0 & 16 & $[16,\,84]$\\
additive, $w=(1,2,3,1,2,3,1,2)$ & 2, 2, 2, 2, 2, 2, 2 & 0, 0, 0, 0, 0, 0, 0 & 56 & $[16,\,56]$\\
multiplicative, $a=(2,3,0,2,3,0,-1,2)$ & 2, 2, 2, 2, 2, 2, 2 & 1, 1, 1, 1, 1, 1, 1 & 56 & $[16,\,104]$\\
pairwise on the path $1-2-\cdots-8$ & 2, 3, 3, 3, 3, 3, 2 & 1, 1, 1, 1, 1, 1, 1 & 104 & $[16,\,104]$\\
threshold $\mathbf 1[|S|\ge4]$ & 2, 3, 4, 5, 4, 3, 2 & 1, 2, 3, 4, 3, 2, 1 & 160 & $[84,\,240]$\\
generic of degree $2$ ($2$-additive) & 2, 4, 5, 6, 5, 4, 2 & 1, 2, 3, 4, 3, 2, 1 & 240 & $[84,\,240]$\\
generic of degree $3$ & 2, 4, 8, 10, 8, 4, 2 & 1, 3, 6, 8, 6, 3, 1 & 488 & $[280,\,488]$\\
generic (Gaussian dividends) & 2, 4, 8, 16, 8, 4, 2 & 1, 3, 7, 15, 7, 3, 1 & 680 & $[520,\,680]$\\
}
\newcommand{\PhiAtt}{\le 10^{-15}}
\newcommand{\RestrMinOne}{1.618}
\newcommand{\RestrMinFive}{1.106}
\newcommand{\RestrMinTen}{1.020}
\newcommand{\RestrMaxOne}{0.618}
\newcommand{\RestrMaxFive}{0.904}
\newcommand{\RestrMaxTen}{0.981}

\newcommand{\SiiN}{300}
\newcommand{\SiiErr}{\le 10^{-14}}
\newcommand{\SiiViol}{0}
\newcommand{\SiiExcess}{4}
\newcommand{\SiiUnaN}{36}
\newcommand{\SiiUnaViol}{0}
\newcommand{\SiiConvViol}{0}
\newcommand{\SiiInvErr}{\le 10^{-14}}
\newcommand{\SiiSharpN}{11}
\newcommand{\SiiSharpViol}{0}
\newcommand{\BernHankelM}{30}

\newcommand{\ExcMax}{7.6\times 10^{-15}}
\newcommand{\ExcSandwich}{1{,}948}
\newcommand{\ExcInterlace}{1{,}248}
\newcommand{\ExcLambda}{243}
\newcommand{\ExcBanzhaf}{41}

\newcommand{\PilotSent}{24}
\newcommand{\PilotNeg}{12}
\newcommand{\PilotLab}{19}
\newcommand{\PilotModels}{6}
\newcommand{\PilotModelsWord}{six}
\newcommand{\PilotCuts}{1{,}584}
\newcommand{\PilotTensors}{144}
\newcommand{\PilotViol}{0}
\newcommand{\PilotEtaMax}{1.8\times10^{-4}}
\newcommand{\PilotAccRange}{16--19}
\newcommand{\PilotRTwoRange}{11.5--17.5}
\newcommand{\PilotRThreeRange}{28--34}
\newcommand{\PilotRMaxAll}{26}
\newcommand{\PilotGamRange}{0.14--0.25}
\newcommand{\PilotSRange}{0.54--0.63}
\newcommand{\PilotRatMax}{0.91}
\newcommand{\PilotRatMedRange}{0.38--0.44}
\newcommand{\PilotDegOneRange}{0.51--0.68}
\newcommand{\PilotDegTwoRange}{0.17--0.23}
\newcommand{\PilotDegThreeRange}{0.14--0.27}

\newcommand{\PilotGamSig}{2}
\newcommand{\PilotDegSig}{2}
\newcommand{\PilotGamPRange}{0.010--0.507}
\newcommand{\PilotDegPRange}{0.006--0.194}
\newcommand{\PilotSPMin}{0.157}
\newcommand{\PilotGamHolm}{0}
\newcommand{\PilotDegHolm}{1}

\newcommand{\PilotNatRange}{11.5--17.5}
\newcommand{\PilotRandRange}{14.4--20.1}
\newcommand{\TablePilot}{Qwen3-1.7B & 17/19 & $1.5\times10^{-4}$ & 13.5 & 31 & 26 & 0.25 & 0.62 & 0.81 & 0.27 & 0.36 / 0.24 & 13.5 / 17.5\\
Qwen3-4B & 19/19 & $8.0\times10^{-5}$ & 12 & 28 & 25 & 0.19 & 0.54 & 0.85 & 0.17 & 0.26 / 0.16 & 12 / 14.9\\
Qwen3-8B & 18/19 & $5.3\times10^{-5}$ & 12.5 & 30 & 21 & 0.20 & 0.54 & 0.87 & 0.18 & 0.28 / 0.17 & 12.5 / 15.5\\
OLMo-2-1B-Instruct & 19/19 & $1.8\times10^{-4}$ & 11.5 & 28 & 18 & 0.14 & 0.58 & 0.91 & 0.14 & 0.16 / 0.14 & 11.5 / 14.4\\
OLMo-2-7B-Instruct & 17/19 & $3.3\times10^{-5}$ & 16 & 34 & 23 & 0.21 & 0.63 & 0.89 & 0.22 & 0.27 / 0.19 & 16 / 19.1\\
Mistral-7B-Instruct & 16/19 & $5.3\times10^{-5}$ & 17.5 & 33 & 24 & 0.23 & 0.62 & 0.80 & 0.22 & 0.30 / 0.21 & 17.5 / 20.1\\}

\title{Cut Rank of Coalitional Value Tensors Counts\\ Cross-Cut Harsanyi Interactions}
\author{Oguzhan Katar\\[2pt]
\small Department of Software Engineering, Faculty of Technology,\\
\small Firat University, Elazig, T\"urkiye\\
\small \href{mailto:okatar@firat.edu.tr}{okatar@firat.edu.tr}}
\date{November 2026}

\begin{document}
\maketitle

\begin{abstract}
Shapley-type explanations of machine learning models, including large language models (LLMs), are computed from a value function on the $2^N$ coalitions of $N$ players, stored as an order-$N$ tensor. In a tensor-train or matrix product state representation, the smallest bond dimension at a cut $L|R$ is the Schmidt rank $\chi$ of the unfolding. We show that $\chi$ counts, to within two, the linearly independent interaction patterns that cross the cut: $\rank C\le\chi\le\rank C+2$, where $C$ is the matrix of Harsanyi dividends of coalitions with players on both sides. For dividends supported on a given family of coalitions the generic $\chi$ is a bipartite matching number, which gives closed forms for games of degree at most $k$. Symmetric, threshold and glove games have $\chi\le\min(|L|,|R|)+1$ although their dividends can be dense, additive games with nonzero weights on both sides have $\chi=2$, and pairwise games are fixed to within two by their cross-cut interaction matrix. As quantum states, nondegenerate additive and multiplicative games are locally equivalent to the W and Greenberger--Horne--Zeilinger states. Numerical ranks obey the sandwich up to factors $\varphi^{\pm N}$, with $\varphi$ the golden ratio, and losslessly in the Walsh--Hadamard (Fourier) basis, where the cross-cut block is a rescaled Banzhaf interaction matrix. Let $\gamma$ be the variance share of the Fourier coefficients that straddle the cut. For $\gamma\le1-2^{1-\min(|L|,|R|)}$ we give the largest R\'enyi-$\alpha$ entropy compatible with $\gamma$ and the cut size. For $\gamma>0$ it is bounded uniformly in $N$ for $\alpha>1$ and unbounded for $\alpha\le1$, and no lower bound in terms of $\gamma$ exists. All bounds were checked on \NInst{} random games and structured families. Exact value tensors of six LLMs on sentiment sentences have moderate ranks, smaller in the word order than in random orders.
\end{abstract}

\section{Introduction}\label{sec:intro}

Shapley-value attributions, such as SHapley Additive exPlanations (SHAP)~\citep{lundberg2017}, and their interaction indices explain a prediction by evaluating a value function $v(S)$ on coalitions $S$ of input features, here the words or tokens of a prompt~\citep{shapley1953,lundberg2017,grabisch1999,sundararajan2020,tsai2023,fumagalli2023}. The object underneath every one of these methods is the full value table, which has $2^N$ entries for $N$ players. Writing $x_i=1$ when player $i$ is present, the table becomes a tensor $\V\in\R^{2\times\cdots\times 2}$ with $\V_{x_1\dots x_N}=v(\{i:x_i=1\})$. Tensor networks are a standard way to store and process such tensors in compressed form. A tensor-train (TT) decomposition~\citep{oseledets2011} writes $\V$ as a product of $N$ small three-way cores, and it is the same object as a matrix product state (MPS) in quantum many-body physics~\citep{schollwock2011,verstraete2008,orus2014,cirac2021}. The smallest TT rank at the cut between players $\{1,\dots,k\}$ and $\{k+1,\dots,N\}$ equals the rank of the corresponding matrix unfolding of $\V$~\citep{oseledets2011}. In physical language it is the Schmidt rank of the normalized tensor read as a bipartite state~\citep{horodecki2009}.

This paper asks what the cut rank says about the interactions in the game. The question arises in three communities, for different reasons.

In explainable artificial intelligence (XAI), exact Shapley values and interaction indices of a large language model (LLM) require all $2^N$ coalition values, so practical methods either sample~\citep{fumagalli2023,fumagalli2024,kolpaczki2024,muschalik2024} or assume structure. Sparse M\"obius and Fourier methods learn a small set of interactions from few queries and have been scaled to LLM inputs~\citep{kang2024,kang2025,butler2025,gorji2025}. Tensor-network methods instead fit a TT or tree surrogate of the value function, from which Shapley values and interaction indices are computed exactly or with certified error~\citep{ballester2022,heidari2026tnshap,heidari2026tnshapg,kim2026ttfsi}, and exact Shapley-type explanations are tractable for models that are themselves TTs~\citep{marzouk2025}. Tensor cross interpolation (TCI) learns such a surrogate from queries with a budget that grows with the TT ranks~\citep{oseledets2010,savostyanov2014,nunez2025}. Whether a surrogate is small therefore depends on the cut ranks of the game. Users of these methods need to know which interaction patterns make the ranks small and which make them large.

In cooperative game theory, interaction structure is classically measured by the M\"obius (Harsanyi) transform, by interaction indices~\citep{grabisch1999,fujimoto2006} and by $k$-additivity, the vanishing of all dividends of coalitions with more than $k$ players~\citep{grabisch1997}. Graph games, whose value is the total weight of the edges inside a coalition, are the standard example of a game with only pairwise interactions~\citep{deng1994}. Cut rank is a different complexity measure. It is invariant under local invertible transformations, it stays at two for games that are dense in the M\"obius basis but multiplicatively separable, and it controls the cost of exact tensor-network computation. Relating it to dividends, $k$-additivity and interaction indices places the measure inside the existing theory.

In tensor networks and quantum information, Schmidt rank and R\'enyi entanglement entropies decide whether a state can be compressed into an MPS~\citep{vidal2003,schuch2008,eisert2010}. Tensor networks have also been used as machine learning models~\citep{stoudenmire2016,han2018}, and entanglement scaling has been used to analyse and choose their architectures~\citep{levine2019,convy2022,lu2025}. Quantum algorithms for Shapley values have been proposed~\citep{burge2023,burge2024}, and Shapley values have been used to explain quantum circuits~\citep{heese2025}. Value tensors of games form a structured class of real states. It is useful to know which known state families they contain and how their entanglement is tied to quantities that can be measured, such as the cross-cut share of the variance.

For coalitional games the relation between cut rank and interactions can be made exact, and the main result is a two-sided bound with additive slack of at most two. The M\"obius transform from values to Harsanyi dividends~\citep{harsanyi1963,grabisch2000} acts as an invertible map on each player separately. The cut rank of $\V$ therefore equals the cut rank of the dividend tensor, which is the invariance of Schmidt rank under local invertible operations~\citep{dur2000,delathauwer2000}. Once the dividend unfolding is split into a boundary part and a cross-cut part, the cut rank is fixed by the rank of the cross-cut dividend matrix to within two. The rest of the paper builds on this observation.

The contributions of the paper are as follows.
\begin{enumerate}[leftmargin=*,itemsep=2pt]
\item Exact cut rank (Theorem~\ref{thm:exact}, Corollary~\ref{cor:exact}). For every cut $[N]=L\sqcup R$, $\rank C_{\cut}\le\chi_{\cut}\le\rank C_{\cut}+2$, where $C_{\cut}$ collects the Harsanyi dividends of coalitions that meet both $L$ and $R$. Multiplicative games have $2^N-1$ nonzero dividends and $\chi=2$ at every cut, so low cut rank does not require sparsity.
\item Generic cut rank of a dividend support (Proposition~\ref{prop:support}). If the dividends vanish outside a family $\mathcal T$ of coalitions, then $\chi$ is at most the matching number of a bipartite graph built from $\mathcal T$, with equality for generic dividends. For games of degree at most $k$, including $k$-additive games, this gives closed forms $\kappa_k$ and $\rho_k$ for $\chi$ and $\rank C$. They are attained generically, and $\rho_k$ is strictly smaller than the bound obtained by grouping rows by size in \DegStrict{} of \DegCasesThirty{} cases with $N\le30$.
\item Structured classes and a dictionary of states (Propositions~\ref{prop:blocksym}--\ref{prop:ttparams}). Games that depend only on $(|S\cap L|,|S\cap R|)$ have $\chi=\rank H$ for a matrix $H$ of size $(|L|+1)\times(|R|+1)$. This gives exact cut ranks for symmetric, threshold, exactly-$w$ and glove games, all of which can have most of their dividends nonzero. Additive games with nonzero weights on both sides of the cut have $\chi=2$, and pairwise games are fixed to within two by the rank of the cross-cut interaction matrix. The minimal TT parameter count is bounded on both sides by the cross-cut ranks. Read as quantum states, these families correspond to product, W-class, Greenberger--Horne--Zeilinger (GHZ)-class and Dicke-space states.
\item Numerical cut rank (Theorem~\ref{thm:numerical}, Proposition~\ref{prop:epsrank}). Thresholded rank and tail-energy rank satisfy the same sandwich up to factors $\varphi^{\pm N}$ and $\varphi^{\pm 4N}$, which come from the conditioning of the M\"obius transform.
\item Fourier, Banzhaf and Shapley interactions (Theorem~\ref{thm:fourier}, Lemma~\ref{lem:banzhaf}, Corollary~\ref{cor:banzhaf}, Proposition~\ref{prop:sii}). In the orthonormal Walsh--Hadamard basis the sandwich holds with no loss, and the cross-cut Fourier block is a diagonal rescaling of the cross-cut Banzhaf interaction matrix. The Shapley interaction index is not a local transform. Its unfolding rank and $\chi$ bound each other up to the factor $\min(|L|,|R|)+1$ in both directions, and both factors are attained, so neither is within an additive constant of the other.
\item R\'enyi entropies (Theorems~\ref{thm:renyi} and~\ref{thm:renyiall}, Corollary~\ref{cor:alphasmall}, Proposition~\ref{prop:nolower}). For the mean-zeroed tensor we give the exact maximum of every R\'enyi entropy $S^{(\alpha)}$ at a given cross-cut share $\gamma\le1-2^{1-\min(|L|,|R|)}$ and cut size, and an upper bound for larger $\gamma$. For $\alpha>1$ the maximum is bounded uniformly in $N$, and $S^{(2)}\le\ln2-2\ln(1-\gamma)$ is the case $\alpha=2$. For $\alpha\le1$ it diverges with the cut size at every fixed $\gamma>0$. Unanimity games rule out every positive nondecreasing lower bound in terms of $\gamma$.
\item Verification and open problems (Sections~\ref{sec:numerics} and~\ref{sec:open}). A self-contained script checks the theorems and propositions on \NInst{} random games and on structured families. Ranks of integer-valued games are computed exactly over a prime field. Six open problems are stated in mathematical form.
\item A pilot on six LLMs (Section~\ref{sec:pilot}). Exact value tensors of \PilotModelsWord{} models from the Qwen3, OLMo 2 and Mistral families on \PilotSent{} sentiment sentences with twelve words have moderate $\varepsilon$-ranks that peak in the middle of the sentence, a share of \PilotDegThreeRange{} of their variance in interactions of order three and higher, and smaller bond dimensions in the word order than in random orders. Sentences with negation tend to show more interaction across the middle cut, but the difference is significant in only \PilotGamSig{} of the \PilotModels{} models before correction for multiple comparisons.
\end{enumerate}

Section~\ref{sec:related} reviews related work and Section~\ref{sec:setup} fixes notation (Table~\ref{tab:notation}). Section~\ref{sec:exact} proves the exact results and Section~\ref{sec:classes} applies them to structured classes of games. Sections~\ref{sec:numerical} and~\ref{sec:fourier} treat numerical ranks in the M\"obius and Fourier bases, and Section~\ref{sec:entropy} treats R\'enyi entropies. Section~\ref{sec:numerics} reports the numerical verification, Section~\ref{sec:pilot} a pilot measurement on six LLMs, Section~\ref{sec:open} lists open problems and Section~\ref{sec:discussion} discusses consequences and limitations. Longer proofs are in the Appendix.

\section{Related work}\label{sec:related}

\subsection{Shapley interaction indices}
The Shapley interaction index (SII) of \citet{grabisch1999} extends the Shapley value~\citep{shapley1953} to coalitions, and the Banzhaf interaction index is its counterpart with uniform weights. Probabilistic and cardinal-probabilistic interaction indices were characterized axiomatically by \citet{fujimoto2006}. The Shapley--Taylor interaction index (STII)~\citep{sundararajan2020} and the Faithful Shapley interaction index (Faith-Shap)~\citep{tsai2023} distribute the value among interactions of bounded order. SHAP-IQ~\citep{fumagalli2023}, KernelSHAP-IQ~\citep{fumagalli2024} and SVARM-IQ~\citep{kolpaczki2024} approximate these indices from sampled coalitions, and the shapiq library collects such estimators~\citep{muschalik2024}. Conversions between the M\"obius transform, interaction indices and other representations of set functions are collected by \citet{grabisch2000}. Our Corollary~\ref{cor:banzhaf} and Proposition~\ref{prop:sii} describe how these conversions act on cut ranks.

\subsection{\texorpdfstring{$k$}{k}-additivity and graph games}
\citet{grabisch1997} introduced $k$-order additive fuzzy measures, whose M\"obius transform vanishes on sets with more than $k$ elements, as a model of bounded interaction. In the weighted graph games of \citet{deng1994} the value of a coalition is the total weight of the edges it contains, so they are $2$-additive. Proposition~\ref{prop:support} gives the generic cut rank of $k$-additive games, and Proposition~\ref{prop:addpair} gives the cut rank of graph games in terms of the cross-cut edges.

\subsection{Tensor networks for Shapley values}
\citet{ballester2022} learns the TT of a cooperative game by cross approximation and evaluates semivalues from it, and \citet{ballester2018} compute advanced sensitivity metrics from a TT that holds all Sobol indices. \citet{marzouk2025} show that exact marginal SHAP values are tractable for TTs, and even in the parallel class NC$^2$, but $\#$P-hard for general tensor networks. \citet{heidari2026tnshap} compute Shapley values and interaction indices from a multilinear tensor-network surrogate (TN-SHAP), and \citet{heidari2026tnshapg} extend this to graph-structured surrogates (TN-SHAP-G). They prove that the rank of the $A|B$ unfolding is at most the product of the bond dimensions on any set of edges whose removal separates $A$ from $B$ (their Theorem~3.2). \citet{kim2026ttfsi} compute Faith-Shap interactions with tensor trains, and \citet{hasegawa2025} write the STII as a linear map with a multilinear weight tensor. \citet{mohammadi2026} compute Shapley values of product games, whose values factorize across players. None of these works bounds the TT ranks of a game by the rank of its cross-cut dividend or interaction matrix. Our Theorem~\ref{thm:exact} complements the separator bound of \citet{heidari2026tnshapg} by tying the rank of an unfolding, to within two, to its cross-cut dividend matrix.

\subsection{Sparse M\"obius and Fourier methods}
Sparse transform methods recover a small set of nonzero dividends or Fourier coefficients from few queries~\citep{kang2024,kang2025,butler2025}, including SPEX and ProxySPEX for LLM inputs, and SHAP values can be computed exactly from a sparse Fourier representation~\citep{gorji2025}. These methods rely on sparsity and low degree. Low cut rank is a different structural assumption (Section~\ref{sec:sparsity}).

\subsection{Tensor networks in machine learning, separation rank and entanglement}
MPS have been used as supervised and generative models~\citep{stoudenmire2016,han2018}, and the scaling of mutual information in classical data has been used to choose tensor-network architectures~\citep{convy2022,lu2025}. Separation rank was used by \citet{razin2023} to quantify the interaction that a graph neural network (GNN) can model across a partition of its input vertices, and they showed that it is governed by the walk index of the partition, a count of the walks that start at its boundary. \citet{levine2019} showed that deep convolutional and recurrent networks can carry volume-law entanglement. These works measure the capacity of an architecture. They do not address the interaction structure of a given coalitional game.

\subsection{Quantum algorithms and explanations}
\citet{burge2023,burge2024} give quantum algorithms that estimate Shapley values, and \citet{heese2025} use Shapley values to explain quantum circuits. R\'enyi-2 entanglement entropy can be estimated by randomized measurements~\citep{brydges2019}. Our dictionary (Table~\ref{tab:dictionary}) identifies which value tensors are related to standard multiqubit states by invertible local operators, in the sense of stochastic local operations and classical communication (SLOCC) equivalence~\citep{dur2000}.

\subsection{Algebraic background}
Fourier analysis on the Boolean cube is treated by \citet{odonnell2014}, and \citet{hammer1992} show that the Banzhaf power indices are the linear coefficients of the best least-squares approximation of a pseudo-Boolean function by one of degree at most one. Owen's multilinear extension expresses the Shapley value as an integral along the diagonal of the cube~\citep{owen1972}, and Proposition~\ref{prop:sii} uses the analogous integral for interactions. Invariance of Schmidt rank under local invertible maps is standard in entanglement theory~\citep{dur2000,horodecki2009}, and invariance of multilinear rank under invertible mode products is standard in multilinear algebra~\citep{delathauwer2000}. Our contribution is to apply these facts to the M\"obius transform of a game and to work out their consequences for interaction indices, structured classes of games and entanglement entropies.

\section{Setting and notation}\label{sec:setup}

Let $[N]=\{1,\dots,N\}$ and let $v:2^{[N]}\to\R$ be a set function (a game). We do not require $v(\varnothing)=0$. The value tensor is
\begin{equation}\label{eq:tensor}
\V\in\R^{2\times\cdots\times 2},\qquad \V_{x_1\dots x_N}=v\bigl(\{i:x_i=1\}\bigr).
\end{equation}
The Harsanyi dividends (M\"obius transform) are $d_T=\sum_{U\subseteq T}(-1)^{|T|-|U|}v(U)$, which invert to $v(S)=\sum_{T\subseteq S}d_T$~\citep{harsanyi1963,grabisch2000}. The dividend tensor is $\D_{\mathbf t}=d_{\{i:t_i=1\}}$. The map $v\mapsto\V$ is a bijection between games and real tensors of format $2\times\cdots\times2$. When $\V\ne0$, the normalized tensor $\V/\fro{\V}$ can therefore be read as a real state vector of $N$ qubits, written in the computational basis $|x_1\dots x_N\rangle$. We use this reading only to borrow terminology. No physical system is involved.

\begin{lemma}[M\"obius transform as a Kronecker product]\label{lem:kron}
Let $Z=\begin{pmatrix}1&0\\1&1\end{pmatrix}$ act on every mode. Then $\V=(Z\otimes\cdots\otimes Z)\,\D$. For a cut $[N]=L\sqcup R$ with $L,R\ne\varnothing$, let $V_{\cut}\in\R^{2^{|L|}\times 2^{|R|}}$ be the unfolding with rows indexed by $x_L$ and columns by $x_R$, and let $D_{\cut}$ be the unfolding of $\D$ indexed in the same way. Then
\begin{equation}\label{eq:unfold}
V_{\cut}=Z^{\otimes|L|}\,D_{\cut}\,\bigl(Z^{\otimes|R|}\bigr)^{\top}.
\end{equation}
\end{lemma}
\begin{proof}
For $x,t\in\{0,1\}^N$ we have $\prod_i Z_{x_it_i}=1$ if $t_i\le x_i$ for all $i$ and $0$ otherwise, because $Z_{01}=0$ and all other entries equal $1$. With $S=\{i:x_i=1\}$ and $T=\{i:t_i=1\}$ this gives $\bigl((Z\otimes\cdots\otimes Z)\D\bigr)_x=\sum_{T\subseteq S}d_T=v(S)$. Splitting the product over $i\in L$ and $i\in R$ gives $\V_{x_Lx_R}=\sum_{t_L,t_R}\bigl(\prod_{i\in L}Z_{x_it_i}\bigr)\D_{t_Lt_R}\bigl(\prod_{i\in R}Z_{x_it_i}\bigr)$. The two products are the entries of $Z^{\otimes|L|}$ and $Z^{\otimes|R|}$ when $x_L,t_L$ and $x_R,t_R$ are ordered consistently, and this is~\eqref{eq:unfold}.
\end{proof}

We order the rows of $D_{\cut}$ by subsets $T_L\subseteq L$ and the columns by $T_R\subseteq R$, with $\varnothing$ first. Then $(D_{\cut})_{T_L,T_R}=d_{T_L\cup T_R}$ and
\begin{equation}\label{eq:blocks}
D_{\cut}=\begin{pmatrix} d_\varnothing & a^{\top}\\ b & C_{\cut}\end{pmatrix},\qquad
C_{\cut}=\bigl(d_{T_L\cup T_R}\bigr)_{T_L\ne\varnothing,\,T_R\ne\varnothing},
\end{equation}
with $a=(d_{T_R})_{\varnothing\ne T_R\subseteq R}$ and $b=(d_{T_L})_{\varnothing\ne T_L\subseteq L}$.
The first row and first column hold the one-sided dividends, meaning those of coalitions contained in $R$ or in $L$. The \emph{cross-cut dividend matrix} $C_{\cut}$, of size $(2^{|L|}-1)\times(2^{|R|}-1)$, holds the dividends of coalitions $T$ with $T\cap L\ne\varnothing$ and $T\cap R\ne\varnothing$: its row is indexed by $T\cap L$ and its column by $T\cap R$. The \emph{cut rank} is $\chi_{\cut}=\rank V_{\cut}$. It equals the smallest TT rank at the cut when $L=\{1,\dots,k\}$~\citep{oseledets2011}, and the Schmidt rank of $\V/\fro{\V}$ across $L|R$~\citep{horodecki2009}.

For a matrix $A\in\R^{m\times n}$ we write $\sigma_1(A)\ge\sigma_2(A)\ge\cdots$ for its singular values, with $\sigma_k(A)=0$ for $k>\min(m,n)$. For $\tau>0$, $\rk{\tau}A=\#\{k:\sigma_k(A)>\tau\}$, and $\eta_k(A)=\sum_{j>k}\sigma_j(A)^2$ is the tail energy. We write $\opn{\cdot}$ for the spectral norm and $\fro{\cdot}$ for the Frobenius norm, and $\varphi=(1+\sqrt5)/2$. Table~\ref{tab:notation} summarizes the notation.

\begin{table}[t]
\centering\small
\caption{Notation.}
\label{tab:notation}
\begin{tabularx}{\linewidth}{@{}lX@{}}
\toprule
symbol & meaning\\
\midrule
$v$, $\V$, $\bar\V$ & game, its value tensor~\eqref{eq:tensor}, and the mean-zeroed tensor (Section~\ref{sec:entropy})\\
$d_T$, $\D$ & Harsanyi dividend of $T$ and the dividend tensor\\
$Z$, $\Hh$ & M\"obius factor $\bigl(\begin{smallmatrix}1&0\\1&1\end{smallmatrix}\bigr)$ and orthonormal Hadamard factor $2^{-1/2}\bigl(\begin{smallmatrix}1&1\\1&-1\end{smallmatrix}\bigr)$\\
$L|R$, $V_{\cut}$, $D_{\cut}$ & cut of $[N]$, unfoldings of $\V$ and $\D$ (rows $L$, columns $R$)\\
$C_{\cut}$, $a$, $b$ & cross-cut dividend block and one-sided dividends, Eq.~\eqref{eq:blocks}\\
$\chi_{\cut}$ & cut rank $\rank V_{\cut}$ (Schmidt rank, smallest TT rank)\\
$\rho_k$, $\kappa_k$ & generic $\rank C$ and $\chi$ of games of degree at most $k$, Eq.~\eqref{eq:rhokappa}\\
$\nu(G)$ & size of a maximum matching of a bipartite graph $G$\\
$\F$, $\hat v$, $\omega_S$ & Fourier tensor, Fourier--Walsh coefficients, Walsh characters, Eq.~\eqref{eq:fourier}\\
$\Cx_{\cut}$ & cross-cut block of the Fourier unfolding\\
$\IB$, $B^\times_{\cut}$; $\ISh$ & Banzhaf interaction index and its cross-cut matrix; Shapley interaction index\\
$\sigma_k$, $\rk\tau$, $\eta_k$, $r_\varepsilon$ & singular values, thresholded rank, tail energy, $\varepsilon$-rank\\
$\lambda_k$, $S^{(\alpha)}$, $\gamma_{\cut}$ & normalized singular values, R\'enyi entropy and cross-cut Fourier share, Eq.~\eqref{eq:renyi}\\
$\Phi_\alpha(\gamma;n)$ & largest $S^{(\alpha)}$ at share $\gamma$ and dimension $n$, Theorem~\ref{thm:renyiall}\\
$\varphi$ & golden ratio $(1+\sqrt5)/2$\\
\bottomrule
\end{tabularx}
\end{table}

\section{Exact cut rank}\label{sec:exact}

\begin{theorem}[Cut rank and cross-cut dividends]\label{thm:exact}
For every cut $[N]=L\sqcup R$ with $L,R\ne\varnothing$,
\begin{equation}\label{eq:thm1}
\rank C_{\cut}\;\le\;\chi_{\cut}\;\le\;\rank C_{\cut}+2 .
\end{equation}
\end{theorem}
\begin{proof}
Since $\det Z=1$, the Kronecker powers $Z^{\otimes|L|}$ and $Z^{\otimes|R|}$ are invertible, and~\eqref{eq:unfold} gives $\chi_{\cut}=\rank D_{\cut}$. The matrix $C_{\cut}$ is a submatrix of $D_{\cut}$, so $\rank C_{\cut}\le\rank D_{\cut}$. For the other side, let $e_1$ be the first standard basis vector. Then
\[
D_{\cut}=\begin{pmatrix}0&0\\0&C_{\cut}\end{pmatrix}+e_1\,(d_\varnothing,a^{\top})+\begin{pmatrix}0\\ b\end{pmatrix}e_1^{\top},
\]
and rank is subadditive, so $\rank D_{\cut}\le\rank C_{\cut}+2$.
\end{proof}

\begin{remark}\label{rem:field}
The proof uses only that $Z$ is invertible and that rank is subadditive, so~\eqref{eq:thm1} holds over every field. Section~\ref{sec:numerics} uses this to check the theorem exactly over $\mathrm{GF}(p)$, the finite field with $p$ elements, for integer-valued games. The identity $\chi=\rank D$ is the invariance of Schmidt rank under local invertible operations~\citep{dur2000}, applied to the M\"obius map. Theorem~\ref{thm:exact} adds the observation that, in the dividend basis, the part of the unfolding that does not straddle the cut has rank at most two.
\end{remark}

For integers $a,b\ge1$ and $k\ge1$ define
\begin{equation}\label{eq:rhokappa}
\rho_k(a,b)=\min_{1\le t\le k}\Bigl(\sum_{j=1}^{t-1}\binom{a}{j}+\sum_{i=1}^{k-t}\binom{b}{i}\Bigr),\qquad
\kappa_k(a,b)=\min_{0\le t\le k+1}\Bigl(\sum_{j=0}^{t-1}\binom{a}{j}+\sum_{i=0}^{k-t}\binom{b}{i}\Bigr),
\end{equation}
with empty sums equal to zero and $\binom{a}{j}=0$ for $j>a$. Taking $t$ optimal for $\rho_k$ in the definition of $\kappa_k$ shows $\kappa_k\le\rho_k+2$. For $k=2$ the choices $t=1,2$ give $\rho_2(a,b)=\min(a,b)$.

\begin{corollary}\label{cor:exact}
Let $L,R\ne\varnothing$.
\begin{enumerate}[label=(\alph*),leftmargin=*,itemsep=1pt]
\item \emph{No cross-cut interaction.} If $d_T=0$ for every $T$ that meets both $L$ and $R$, then $\chi_{\cut}\le 2$.
\item \emph{Low degree.} If $d_T=0$ whenever $|T|>k$ for some $k\ge2$, then $\rank C_{\cut}\le\rho_k(|L|,|R|)$ and $\chi_{\cut}\le\kappa_k(|L|,|R|)\le\rho_k(|L|,|R|)+2$. In particular, when only pairwise interactions are present, $\chi_{\cut}\le 2+\min(|L|,|R|)$.
\item \emph{Sparsity.} If $s$ dividends are nonzero, of which $s_\times$ are cross-cut, then $\chi_{\cut}\le\min(s,\,2+s_\times)$.
\item \emph{Unanimity games.} For $u_T(S)=\one[T\subseteq S]$ we have $\chi_{\cut}=1$ for every cut.
\item \emph{Multiplicative games.} For $v(S)=\prod_{i\in S}a_i-1$ with all $a_i\ne1$, all $2^N-1$ dividends $d_T$ with $T\ne\varnothing$ are nonzero, $\rank C_{\cut}=1$, and $\chi_{\cut}=2$ for every cut.
\end{enumerate}
\end{corollary}
\begin{proof}
(a) Here $C_{\cut}=0$. (b) The entry $(T_L,T_R)$ of $C_{\cut}$ can be nonzero only if $|T_L|+|T_R|\le k$. Fix $t\in\{1,\dots,k\}$. Every nonzero entry has $|T_L|\le t-1$ or $|T_R|\le k-t$, since otherwise $|T_L|+|T_R|\ge k+1$. Hence $C_{\cut}$ is the sum of the matrix formed by its rows with $|T_L|\le t-1$ and a matrix supported on its columns with $|T_R|\le k-t$, and its rank is at most the number of those rows plus the number of those columns. Minimizing over $t$ gives $\rho_k$. The same argument applied to $D_{\cut}$, with $0\le t\le k+1$ and the empty set included, gives $\chi_{\cut}=\rank D_{\cut}\le\kappa_k$. (c) Rank is at most the number of nonzero entries. Apply this to $D_{\cut}$, and to $C_{\cut}$ in~\eqref{eq:thm1}. (d) $\D$ has exactly one nonzero entry. (e) Since $\prod_{i\in S}a_i=\prod_{i\in S}\bigl(1+(a_i-1)\bigr)=\sum_{T\subseteq S}\prod_{i\in T}(a_i-1)$, we get $d_T=\prod_{i\in T}(a_i-1)\ne0$ for $T\ne\varnothing$ and $d_\varnothing=0$. Hence $C_{\cut}=c_Lc_R^{\top}$ with $(c_L)_{T_L}=\prod_{i\in T_L}(a_i-1)$, which has rank one. Moreover $V_{\cut}=\alpha_L\alpha_R^{\top}-\one\one^{\top}$ with $\alpha_L=\bigotimes_{i\in L}(1,a_i)^{\top}$. The vector $\alpha_L$ has entries $1$ and $a_i\ne1$, so it is not parallel to $\one$, and the same holds for $\alpha_R$. A sum of two rank-one terms whose left factors are linearly independent and whose right factors are linearly independent has rank two.
\end{proof}

\begin{figure}[tbp]
\centering
\includegraphics[width=0.96\linewidth]{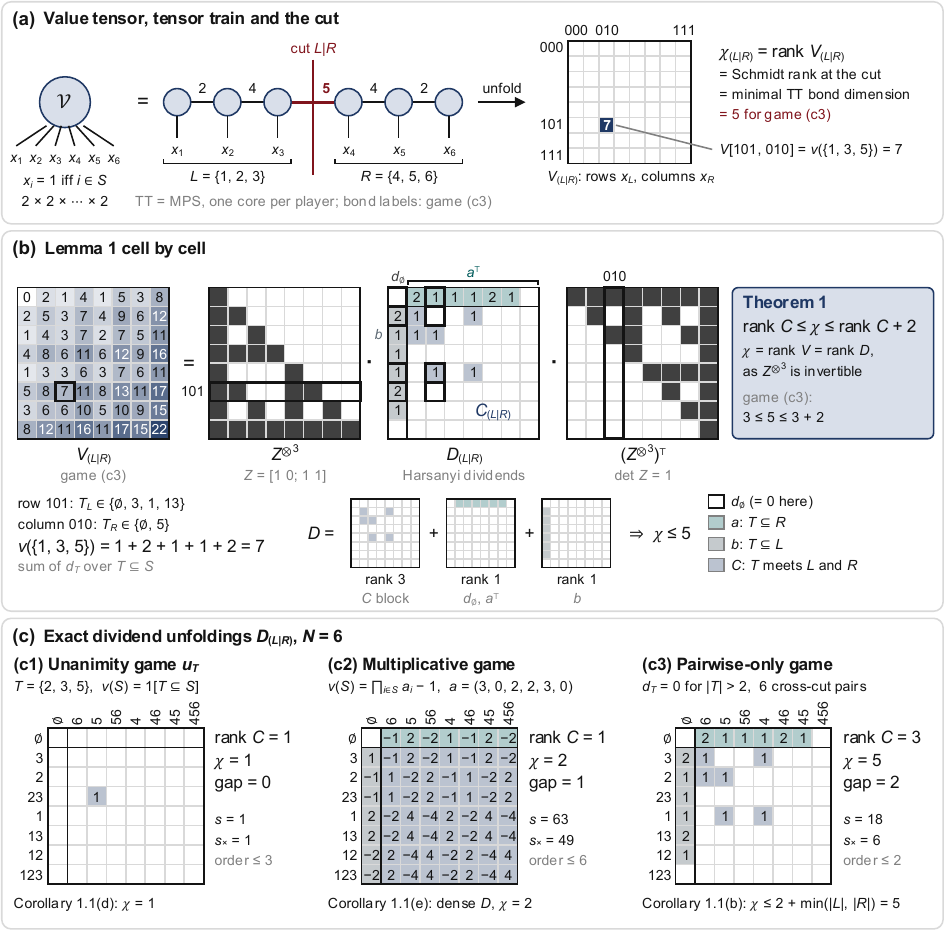}
\caption{Cut rank and cross-cut dividends for $N=6$ and the cut $L|R$ with $L=\{1,2,3\}$ and $R=\{4,5,6\}$. (a)~The value tensor $\V$, its tensor-train (TT) or matrix product state (MPS) representation and the unfolding $V_{\cut}$. The bond across the cut has dimension $\chi_{\cut}=\rank V_{\cut}$, the Schmidt rank of the unfolding; bond labels are the minimal TT ranks of game (c3). (b)~Lemma~\ref{lem:kron} cell by cell for game (c3). Row $101$ of $Z^{\otimes3}$ and column $010$ of $(Z^{\otimes3})^{\top}$ select the outlined dividends $d_T$ with $T\subseteq\{1,3,5\}$, whose sum is the highlighted entry $v(\{1,3,5\})=7$. Splitting $D_{\cut}$ into $C_{\cut}$ and a boundary part of rank at most two gives Theorem~\ref{thm:exact}. (c)~Exact dividend unfoldings $D_{\cut}$; blank cells are zero, rows and columns are in binary order with the empty set first, and $s$ and $s_\times$ count the nonzero and the cross-cut dividends. (c1)~Unanimity game $u_T$ with $T=\{2,3,5\}$: $\rank C=\chi=1$ (Corollary~\ref{cor:exact}(d)). (c2)~Multiplicative game $v(S)=\prod_{i\in S}a_i-1$ with $a=(3,0,2,2,3,0)$: all $63$ dividends with $T\ne\varnothing$ are nonzero, yet $\rank C=1$ and $\chi=2$ (Corollary~\ref{cor:exact}(e)). (c3)~Singleton and pairwise dividends only, $18$ of them nonzero and six cross-cut: $\rank C=3$ and $\chi=5=2+\min(|L|,|R|)$, so Corollary~\ref{cor:exact}(b) is attained. The three games realize the gaps $0$, $1$ and $2$ allowed by Theorem~\ref{thm:exact}.}
\label{fig:cut}
\end{figure}

Figure~\ref{fig:cut} illustrates Lemma~\ref{lem:kron}, the block structure~\eqref{eq:blocks} and Theorem~\ref{thm:exact} on three games with $N=6$. All three values $\chi-\rank C\in\{0,1,2\}$ occur (Section~\ref{sec:numerics}). Theorem~\ref{thm:exact} therefore cannot be improved without further assumptions. The multiplicative game shows that low cut rank does not imply sparsity of the dividends, a point we return to in Section~\ref{sec:sparsity}.

The bounds of Corollary~\ref{cor:exact}(b) are attained by generic dividends. This follows from a general statement about prescribed dividend supports. For a family $\mathcal T\subseteq2^{[N]}$ and a cut $L|R$, let $G_{\mathcal T}$ be the bipartite graph with vertex classes $\{T_L\subseteq L\}$ and $\{T_R\subseteq R\}$ and an edge $\{T_L,T_R\}$ whenever $T_L\cup T_R\in\mathcal T$, and let $G^\times_{\mathcal T}$ be its restriction to nonempty $T_L$ and $T_R$. Thus $G_{\mathcal T}$ is the support graph of $D_{\cut}$ and $G^\times_{\mathcal T}$ that of $C_{\cut}$ when the dividends vanish outside $\mathcal T$. We write $\nu(G)$ for the size of a maximum matching.

\begin{proposition}[Generic cut rank of a dividend support]\label{prop:support}
Let $\mathcal T\subseteq2^{[N]}$, let $L,R\ne\varnothing$, and consider games with $d_T=0$ for $T\notin\mathcal T$.
\begin{enumerate}[label=(\roman*),leftmargin=*,itemsep=1pt]
\item For every such game, $\chi_{\cut}\le\nu(G_{\mathcal T})$ and $\rank C_{\cut}\le\nu(G^\times_{\mathcal T})$.
\item If the dividends $(d_T)_{T\in\mathcal T}$ have a joint distribution that is absolutely continuous with respect to Lebesgue measure on $\R^{\mathcal T}$, then both inequalities in (i) are equalities with probability one.
\item For $\mathcal T=\{T:|T|\le k\}$ with $k\ge2$, $\nu(G_{\mathcal T})=\kappa_k(|L|,|R|)$ and $\nu(G^\times_{\mathcal T})=\rho_k(|L|,|R|)$. Moreover
\begin{equation}\label{eq:grouped}
\begin{aligned}
\rho_k(a,b)&\le\min\Bigl\{\sum_{j=1}^{k-1}\min\Bigl(\binom{a}{j},\sum_{i=1}^{k-j}\binom{b}{i}\Bigr),\;\sum_{j=1}^{k-1}\min\Bigl(\binom{b}{j},\sum_{i=1}^{k-j}\binom{a}{i}\Bigr)\Bigr\}\\
&\le\min\Bigl\{\sum_{j=1}^{k-1}\binom{a}{j},\;\sum_{j=1}^{k-1}\binom{b}{j}\Bigr\}.
\end{aligned}
\end{equation}
\end{enumerate}
\end{proposition}
\begin{proof}[Proof of (i) and (ii)]
(i) Let $M$ be a matrix whose support graph has matching number $\nu$. Every square submatrix of size larger than $\nu$ has zero determinant, because each nonzero term of the Leibniz expansion would be a matching of that size in the support graph. Hence $\rank M\le\nu$. Apply this to $D_{\cut}$ and $C_{\cut}$ and use $\chi_{\cut}=\rank D_{\cut}$. (ii) Each dividend $d_T$ appears in exactly one cell of $D_{\cut}$, namely $(T\cap L,T\cap R)$. Take a maximum matching and the square submatrix on its rows and columns. Its determinant is a polynomial in the dividends. The monomial formed by the matched entries arises from exactly one permutation, because any permutation that produces it must use the same cells, so its coefficient is $\pm1$ and the polynomial is not identically zero. The zero set of a nonzero polynomial has Lebesgue measure zero, so with probability one this minor is nonzero and the rank is at least $\nu$. The same argument applies to $C_{\cut}$.
\end{proof}
Part (iii) is proved in Appendix~\ref{app:support}, through an exact formula for the matching number of bipartite graphs with nested neighbourhoods (Lemma~\ref{lem:nested}). The middle expression in~\eqref{eq:grouped} groups the rows of $C_{\cut}$ by $|T_L|$ and bounds the rank of each group separately. It is a valid bound, but it is not attained in general. Among the \DegCasesThirty{} triples $(|L|,|R|,k)$ with $N\le30$ it exceeds $\rho_k$ in \DegStrict{} cases, the first being $|L|=|R|=\DegFirstL$ and $k=\DegFirstK$, where the generic rank is $\rho_k=\DegFirstRho$, the grouped bound is $\DegFirstGrouped$ and the row count on the right of~\eqref{eq:grouped} is $\DegFirstRow$. Part (i) also contains Corollary~\ref{cor:exact}(c), since every edge of $G_{\mathcal T}$ corresponds to a different coalition, so $\nu(G_{\mathcal T})\le|\mathcal T|$.

\begin{remark}[$k$-additive games]\label{rem:kadd}
A game is $k$-additive in the sense of \citet{grabisch1997} if $d_T=0$ whenever $|T|>k$ and $d_T\ne0$ for some $|T|=k$. Corollary~\ref{cor:exact}(b) and Proposition~\ref{prop:support}(iii) therefore give the cut ranks of $k$-additive games: at most $\kappa_k(|L|,|R|)$, with equality for generic dividends of degree at most $k$. Since $\rho_k(a,b)\le\sum_{j=1}^{k-1}\binom{\min(a,b)}{j}$, the cut rank of a $k$-additive game is $O\bigl(\min(|L|,|R|)^{k-1}\bigr)$ for fixed $k$. Pairwise games, the case $k=2$, are treated in more detail in Proposition~\ref{prop:addpair}.
\end{remark}

\section{Structured classes of games}\label{sec:classes}

This section applies Theorem~\ref{thm:exact} and Proposition~\ref{prop:support} to classes of games that appear in cooperative game theory and in explanation methods, and relates the resulting rank profiles to the minimal size of a TT and to standard multiqubit states.

\begin{proposition}[Block-symmetric games]\label{prop:blocksym}
Let $L|R$ be a cut and suppose that $v(S)=h(|S\cap L|,|S\cap R|)$ for a function $h$ on $\{0,\dots,|L|\}\times\{0,\dots,|R|\}$. Let $H=\bigl(h(s,t)\bigr)$ be the corresponding $(|L|+1)\times(|R|+1)$ matrix. Then $\chi_{\cut}=\rank H\le\min(|L|,|R|)+1$. In particular:
\begin{enumerate}[label=(\alph*),leftmargin=*,itemsep=1pt]
\item \emph{Symmetric games.} If $v(S)=g(|S|)$, then at every cut $\chi_{\cut}$ is the rank of the Hankel matrix $\bigl(g(s+t)\bigr)_{0\le s\le|L|,\,0\le t\le|R|}$, and $\chi_{\cut}=\min(|L|,|R|)+1$ at every cut for generic $g\in\R^{N+1}$.
\item \emph{Threshold games.} For $v_q(S)=\one[|S|\ge q]$ with $0\le q\le N$, $\chi_{\cut}=1+\min(q,N-q,|L|,|R|)$ at every cut. For $q\ge1$ the dividends are $d_T=(-1)^{|T|-q}\binom{|T|-1}{q-1}$ if $|T|\ge q$ and $d_T=0$ otherwise, so $\sum_{t=q}^N\binom{N}{t}$ of them are nonzero.
\item \emph{Exactly-$w$ games.} For $v(S)=\one[|S|=w]$ with $0\le w\le N$, $\chi_{\cut}=1+\min(w,N-w,|L|,|R|)$ at every cut.
\item \emph{Glove games.} If the players in $L$ hold left gloves and those in $R$ right gloves, and $v(S)=\min(|S\cap L|,|S\cap R|)$ is the number of pairs that $S$ can form, then $\chi_{\cut}=\min(|L|,|R|)$. Its dividend $d_T$ is nonzero exactly when $T$ meets both $L$ and $R$, so $(2^{|L|}-1)(2^{|R|}-1)$ dividends are nonzero.
\end{enumerate}
\end{proposition}
\begin{proof}[Proof of the main statement]
Let $A_L\in\{0,1\}^{2^{|L|}\times(|L|+1)}$ have entries $(A_L)_{x_L,s}=\one[|x_L|=s]$, and define $A_R$ in the same way. Then $V_{\cut}=A_LHA_R^{\top}$. The columns of $A_L$ have disjoint nonempty supports, so $A_L$ has full column rank and a left inverse, and the same holds for $A_R$. Hence $\rank V_{\cut}=\rank H$.
\end{proof}
The items (a)--(d) are proved in Appendix~\ref{app:blocksym}. The verification script checks them on all contiguous cuts of every threshold and exactly-$w$ game with $N=6$ to $12$ (\ThrN{} game--cut pairs each; by symmetry these cover every cut size) and on \GloveN{} glove games. The classes are dense in the M\"obius basis. The majority game $\one[|S|\ge\MajQ]$ with $N=\MajN$ has \MajNnz{} nonzero dividends out of \MajTot, while its cut ranks along the natural order are $\MajProfile$. The glove game with $|L|=|R|=6$ has \GloveSixNnz{} nonzero dividends and $\chi_{\cut}=\GloveSixChi$. Methods that assume a sparse M\"obius spectrum do not apply to these games, whereas their TT ranks grow at most linearly in $N$.

\begin{proposition}[Additive and pairwise games]\label{prop:addpair}
Let $L,R\ne\varnothing$.
\begin{enumerate}[label=(\alph*),leftmargin=*,itemsep=1pt]
\item \emph{Additive games.} Let $v(S)=c+\sum_{i\in S}w_i$. If $w_i\ne0$ for some $i\in L$ and $w_j\ne0$ for some $j\in R$, then $\chi_{\cut}=2$. Otherwise $\chi_{\cut}\le1$.
\item \emph{Pairwise games.} Let $v(S)=c+\sum_{i\in S}a_i+\sum_{\{i,j\}\subseteq S}w_{ij}$, and let $W_{LR}=(w_{ij})_{i\in L,\,j\in R}$ be the cross-cut block of the weighted interaction graph. Then $\rank C_{\cut}=\rank W_{LR}$ and
\[
\rank W_{LR}\;\le\;\chi_{\cut}\;\le\;\rank W_{LR}+2 .
\]
Moreover $\rank W_{LR}\le\nu(G_{LR})$, where $G_{LR}$ is the bipartite graph of the cross-cut edges $\{i,j\}$ with $w_{ij}\ne0$, with equality for generic weights on a fixed graph.
\end{enumerate}
\end{proposition}
\begin{proof}
(a) Let $\alpha_L(x_L)=\sum_{i\in L}w_ix_i$ and $\beta_R(x_R)=\sum_{j\in R}w_jx_j$. Then $V_{\cut}=(\alpha_L+c\one)\one^{\top}+\one\beta_R^{\top}$, which has rank at most two. Since $\alpha_L(\varnothing)=0$, the vector $\alpha_L+c\one$ is parallel to $\one$ exactly when $\alpha_L=0$, that is, when $w_i=0$ for all $i\in L$, and similarly $\beta_R$ is parallel to $\one$ exactly when $\beta_R=0$. If both sides carry a nonzero weight, the left factors $\alpha_L+c\one,\one$ and the right factors $\one,\beta_R$ are linearly independent pairs, so the rank is two. If $\beta_R=0$, then $V_{\cut}=(\alpha_L+c\one)\one^{\top}$ has rank at most one, and the case $\alpha_L=0$ is symmetric. (b) The only cross-cut dividends are $d_{\{i,j\}}=w_{ij}$ with $i\in L$, $j\in R$. They sit in the rows $T_L=\{i\}$ and columns $T_R=\{j\}$ of $C_{\cut}$, which is zero elsewhere. So $C_{\cut}$ is $W_{LR}$ padded with zeros, and Theorem~\ref{thm:exact} gives the sandwich. The last statement follows from the argument of Proposition~\ref{prop:support}(i, ii) applied to the matrix $W_{LR}$, whose nonzero entries are distinct weights.
\end{proof}

Proposition~\ref{prop:addpair}(b) applies to the weighted graph games of \citet{deng1994}, which have $c=0$ and $a_i=0$. If the interaction graph is a path $1-2-\dots-N$, every contiguous cut is crossed by one edge, so $\chi\le3$. If it is a star, $\nu(G_{LR})\le1$ at every cut and $\chi\le3$. In general $\chi_{\cut}$ is at most two plus the smallest number of vertices that cover the cross-cut edges. For $2$-additive games with arbitrary weights, $\rank W_{LR}\le\min(|L|,|R|)$, which is the case $k=2$ of Corollary~\ref{cor:exact}(b).

\begin{proposition}[TT parameter count]\label{prop:ttparams}
Order the players naturally, let $\chi_k$ be the cut rank of $\{1,\dots,k\}|\{k+1,\dots,N\}$ and $c_k=\rank C_{(\{1,\dots,k\}|\{k+1,\dots,N\})}$ for $1\le k\le N-1$, and put $\chi_0=\chi_N=1$. A TT with ranks $r_1,\dots,r_{N-1}$ has $\sum_{k=1}^N2r_{k-1}r_k$ parameters, with $r_0=r_N=1$. The smallest number of parameters of a TT that represents $\V\ne0$ exactly is
\[
P_{\min}(v)=\sum_{k=1}^{N}2\chi_{k-1}\chi_k,
\qquad\text{and}\qquad
\sum_{k=1}^{N}2\ell_{k-1}\ell_k\;\le\;P_{\min}(v)\;\le\;\sum_{k=1}^{N}2u_{k-1}u_k,
\]
where $\ell_k=\max(1,c_k)$, $u_k=\min(c_k+2,2^k,2^{N-k})$ for $1\le k\le N-1$, and $\ell_0=\ell_N=u_0=u_N=1$.
\end{proposition}
\begin{proof}
Every TT representation has $r_k\ge\rank V_{(\{1,\dots,k\}|\{k+1,\dots,N\})}=\chi_k$, and the TT singular value decomposition (TT-SVD) attains $r_k=\chi_k$ for all $k$ simultaneously~\citep{oseledets2011}. The parameter count is nondecreasing in every $r_k$, which gives the formula for $P_{\min}$. For $\V\ne0$ we have $\chi_k\ge1$, and Theorem~\ref{thm:exact} gives $c_k\le\chi_k\le c_k+2$, while $\chi_k\le\min(2^k,2^{N-k})$ by the size of the unfolding. Substituting these bounds term by term gives the two inequalities.
\end{proof}

\begin{figure}[tbp]
\centering
\includegraphics[width=\linewidth]{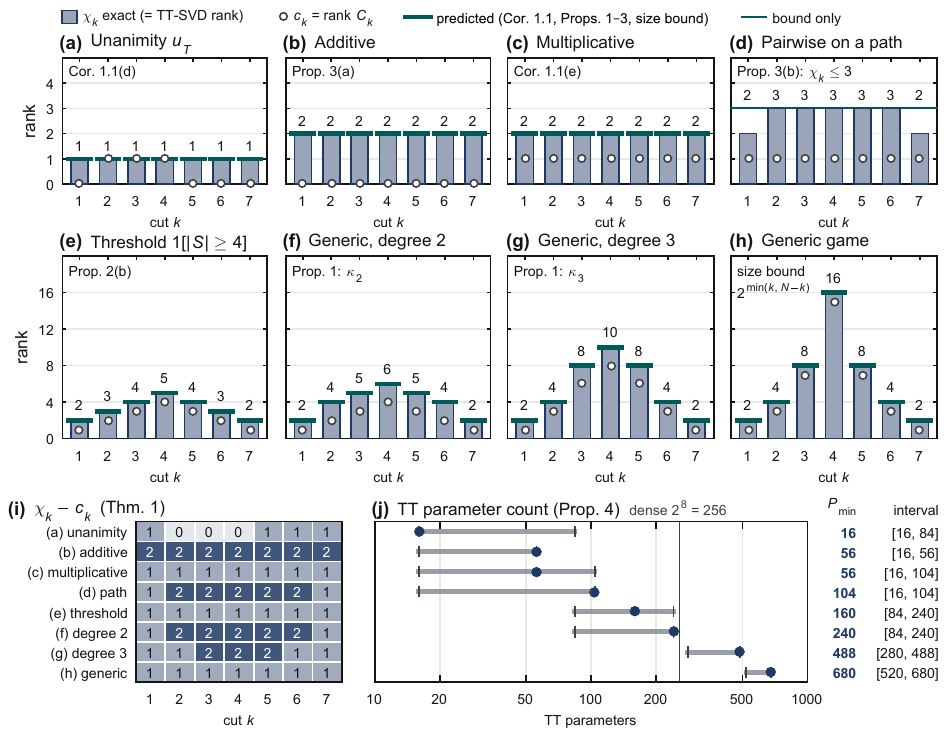}
\caption{Cut-rank profiles $\chi_k$ along the natural player order for eight games with $N=8$, computed exactly (over $\mathrm{GF}(p)$ for integer-valued games, by singular value decomposition for the generic ones; Table~\ref{tab:profiles}). (a)--(h)~Each bar is the cut $\{1,\dots,k\}|\{k+1,\dots,N\}$, and its label is $\chi_k$; TT-SVD returns the same ranks at all $56$ cuts. Thick horizontal marks show the value predicted by Corollary~\ref{cor:exact}, Propositions~\ref{prop:support}--\ref{prop:addpair} and, for the generic game, the size bound $2^{\min(k,N-k)}$; the source is named in each panel. For the path game only the bound $\chi_k\le3$ is predicted (thin line). Open circles mark $c_k=\rank C_k$, the rank of the cross-cut dividend block. The two rows use different vertical scales. (i)~The gap $\chi_k-c_k$, which Theorem~\ref{thm:exact} confines to $\{0,1,2\}$; the values $0$, $1$ and $2$ occur at $3$, $33$ and $20$ of the $56$ cuts. (j)~The smallest TT parameter count $P_{\min}$ of Proposition~\ref{prop:ttparams} (dots, values on the right: $16$, $56$, $56$, $104$, $160$, $240$, $488$, $680$) inside the interval of that proposition (bars). $P_{\min}$ attains the upper end for five families and the lower end for the unanimity game; the vertical line marks the $2^N=256$ entries of the dense tensor.}
\label{fig:profiles}
\end{figure}

For a unanimity game $P_{\min}=2N$. For a multiplicative game, or an additive game with all $w_i\ne0$, $\chi_k=2$ for $1\le k\le N-1$ and $P_{\min}=8N-8$. For a pairwise game on the path with nonzero edge weights, $u_1=u_{N-1}=2$ and $u_k=3$ otherwise, so $P_{\min}\le18N-40$ for $N\ge4$. For a symmetric game $P_{\min}\le\sum_k2(m_{k-1}+1)(m_k+1)$ with $m_k=\min(k,N-k)$, which is $O(N^3)$. For a generic game of degree at most $k$, $\chi_j=\kappa_k(j,N-j)=O(N^{k-1})$ by Remark~\ref{rem:kadd}, so $P_{\min}=O(N^{2k-1})$ for fixed $k$, while a generic game has $P_{\min}\ge2^{N/2}$. Figure~\ref{fig:profiles} and Table~\ref{tab:profiles} show these profiles and $P_{\min}$ for eight games with $N=8$.

\begin{table}[t]
\centering\small
\caption{Game families and the multiqubit states obtained by reading $\V/\fro{\V}$ as a state vector (Section~\ref{sec:setup}). ``SLOCC'' means that the two tensors are related by an invertible operator acting on each qubit separately~\citep{dur2000}; such tensors have the same Schmidt rank at every cut, but they are different states. $|{+}\rangle=(|0\rangle+|1\rangle)/\sqrt2$, $|W\rangle=N^{-1/2}\sum_i|0\cdots1_i\cdots0\rangle$, $|\mathrm{GHZ}\rangle=(|0\cdots0\rangle+|1\cdots1\rangle)/\sqrt2$, and $|D^N_w\rangle$ is the normalized uniform superposition of the basis states of Hamming weight $w$~\citep{bartschi2019}.}
\label{tab:dictionary}
\begin{tabularx}{\linewidth}{@{}>{\raggedright\arraybackslash}p{3.5cm}>{\raggedright\arraybackslash}X>{\raggedright\arraybackslash}p{3.6cm}l@{}}
\toprule
game & value tensor read as a state & cut rank $\chi_{\cut}$ & source\\
\midrule
unanimity $u_T$ & product state $\bigotimes_{i\in T}|1\rangle\otimes\bigotimes_{i\notin T}|{+}\rangle$ & $1$ at every cut & Cor.~\ref{cor:exact}(d)\\[2pt]
$\one[|S|=1]$ & equal to $|W\rangle$ & $2$ at every cut & Prop.~\ref{prop:blocksym}(c)\\[2pt]
additive, all $w_i\ne0$ (shown for $c=0$) & SLOCC-equivalent to $|W\rangle$: $\V=\sqrt N\bigotimes_i\bigl(Z\operatorname{diag}(1,w_i)\bigr)|W\rangle$ & $2$ at every cut & Prop.~\ref{prop:addpair}(a)\\[2pt]
multiplicative \mbox{$\prod_{i\in S}a_i-1$}, all $a_i\ne1$ & $v+1$ is a product state; $v$ is SLOCC-equivalent to $|\mathrm{GHZ}\rangle$ & $2$ at every cut & Cor.~\ref{cor:exact}(e)\\[2pt]
\mbox{$\one[S=\varnothing]+\one[S=[N]]$} & equal to $|\mathrm{GHZ}\rangle$ & $2$ at every cut & direct\\[2pt]
symmetric $g(|S|)$ & permutation invariant, in the span of $|D^N_0\rangle,\dots,|D^N_N\rangle$ & $\le\min(|L|,|R|)+1$ & Prop.~\ref{prop:blocksym}(a)\\[2pt]
exactly-$w$ $\one[|S|=w]$ & equal to $|D^N_w\rangle$ & $1+\min(w,N-w,\allowbreak|L|,|R|)$ & Prop.~\ref{prop:blocksym}(c)\\[2pt]
pairwise ($2$-additive) & no standard name & from $\rank W_{LR}$ to $\rank W_{LR}+2$ & Prop.~\ref{prop:addpair}(b)\\[2pt]
degree $\le k$ ($k$-additive) & no standard name & $\le\kappa_k(|L|,|R|)$, equal generically & Prop.~\ref{prop:support}\\[2pt]
general & every real $N$-qubit state & $\le2^{\min(|L|,|R|)}$, equal generically & Sec.~\ref{sec:setup}\\
\bottomrule
\end{tabularx}
\end{table}

Table~\ref{tab:dictionary} lists the correspondences between game families and multiqubit states. The value tensor of $u_T$ is $\bigotimes_{i\in T}e_1\otimes\bigotimes_{i\notin T}(e_0+e_1)$, a product. For an additive game with $c=0$, $d_{\{i\}}=w_i$ and all other dividends vanish, so $\D=\bigotimes_i\operatorname{diag}(1,w_i)\sum_ie^{(i)}$, where $e^{(i)}$ has a one in position $i$ and zeros elsewhere. Since $\sum_ie^{(i)}=\sqrt N|W\rangle$, Lemma~\ref{lem:kron} gives the entry of Table~\ref{tab:dictionary}, and the local operators are invertible exactly when all $w_i\ne0$. For $c\ne0$ the dividend tensor gains the term $c\,e_0^{\otimes N}$, which the invertible operator $|1\rangle\mapsto|1\rangle-(c/w_1)|0\rangle$ on the first qubit removes, so the game is still SLOCC-equivalent to $|W\rangle$. For a multiplicative game, $\V=\bigotimes_i\alpha_i-\bigotimes_i\one$ with $\alpha_i=(1,a_i)^{\top}$. If $a_i\ne1$, the matrix $M_i=[\one\ \alpha_i]^{-1}$ exists and maps $\one\mapsto e_0$, $\alpha_i\mapsto e_1$, so $\bigotimes_iM_i\,\V=e_1^{\otimes N}-e_0^{\otimes N}$, which is $\sqrt2|\mathrm{GHZ}\rangle$ after the local sign change $\operatorname{diag}(-1,1)$ on one qubit. The value tensor of a symmetric game is $\sum_wg(w)\sqrt{\binom Nw}\,|D^N_w\rangle$, and the cut rank bound in Table~\ref{tab:dictionary} is Proposition~\ref{prop:blocksym}(a). The Schmidt rank of $|D^N_w\rangle$, $1+\min(w,N-w,|L|,|R|)$, follows from Proposition~\ref{prop:blocksym}(c) without appeal to the physics literature. The verification script checks both SLOCC identities numerically for $N=8$ (maximum entrywise error $\SloccErr$).

\begin{table}[t]
\centering\small
\caption{Rank profiles at $N=8$ along the natural order (Figure~\ref{fig:profiles}). Columns give $\chi_k$ and $c_k=\rank C_k$ for $k=1,\dots,7$, the smallest TT parameter count $P_{\min}$, and the interval of Proposition~\ref{prop:ttparams}. TT-SVD reproduced every tensor with the ranks $\chi_k$ (\ProfViol{} discrepancies).}
\label{tab:profiles}
\begin{tabular}{lcccc}
\toprule
game & $\chi_1,\dots,\chi_7$ & $c_1,\dots,c_7$ & $P_{\min}$ & Prop.~\ref{prop:ttparams}\\
\midrule
\TableProfiles
\bottomrule
\end{tabular}
\end{table}

\section{Numerical cut rank in the M\"obius basis}\label{sec:numerical}

In floating point arithmetic exact rank is almost always full, so measurements use thresholded singular values. We first collect the standard singular-value facts we need and prove them from the Eckart--Young--Mirsky theorem.

\begin{lemma}\label{lem:sv}
Let $A,P,Q$ be real matrices of compatible sizes and let $k\ge1$; the statements about $\eta_k$ in (i)--(iv) also hold for $k=0$, where $\eta_0(A)=\fro{A}^2$.
\begin{enumerate}[label=(\roman*),leftmargin=*,itemsep=1pt]
\item \emph{Eckart--Young--Mirsky.} $\sigma_{k}(A)=\min_{\rank X\le k-1}\opn{A-X}$ and $\eta_{k}(A)=\min_{\rank X\le k}\fro{A-X}^2$~\citep{eckart1936,mirsky1960,horn2012}.
\item $\sigma_k(PAQ)\le\opn{P}\opn{Q}\,\sigma_k(A)$ and $\eta_k(PAQ)\le\opn{P}^2\opn{Q}^2\,\eta_k(A)$.
\item If $P,Q$ are square and invertible, then $\sigma_k(PAQ)\ge\sigma_k(A)/(\opn{P^{-1}}\opn{Q^{-1}})$, and the analogous bound holds for $\eta_k$ with squared norms.
\item If $C$ is a submatrix of $D$, then $\sigma_k(C)\le\sigma_k(D)$ and $\eta_k(C)\le\eta_k(D)$.
\item If $C$ is obtained from $D$ by deleting one row and one column, then $\sigma_{k+2}(D)\le\sigma_k(C)$ and $\eta_{k+2}(D)\le\eta_k(C)$. The second inequality also holds for $k=0$.
\end{enumerate}
\end{lemma}
\begin{proof}
(ii) Let $X$ attain the minimum in (i) for $\sigma_k(A)$. Then $\rank(PXQ)\le k-1$, so $\sigma_k(PAQ)\le\opn{P(A-X)Q}\le\opn{P}\opn{Q}\sigma_k(A)$. The tail-energy version follows in the same way from $\fro{PYQ}\le\opn{P}\fro{Y}\opn{Q}$. (iii) Apply (ii) to $A=P^{-1}(PAQ)Q^{-1}$. (iv) $C=S_1DS_2$, where $S_1,S_2$ select rows and columns and have spectral norm at most one. Apply (ii). (v) Permute the deleted row and column to the first position, so that $D=\bigl(\begin{smallmatrix}\delta&r^{\top}\\ c&C\end{smallmatrix}\bigr)$. Let $X$ be a best rank-$(k-1)$ approximation of $C$ in the spectral norm, or a best rank-$k$ approximation in the Frobenius norm. Set $Y=\bigl(\begin{smallmatrix}\delta&r^{\top}\\ c&X\end{smallmatrix}\bigr)=\bigl(\begin{smallmatrix}0&0\\0&X\end{smallmatrix}\bigr)+e_1(\delta,r^{\top})+\bigl(\begin{smallmatrix}0\\ c\end{smallmatrix}\bigr)e_1^{\top}$. Then $\rank Y\le\rank X+2$ and $D-Y=\bigl(\begin{smallmatrix}0&0\\0&C-X\end{smallmatrix}\bigr)$, and (i) gives both inequalities. For $k=0$ in the Frobenius case take $X=0$.
\end{proof}

\begin{lemma}\label{lem:Z}
The singular values of $Z$ are $\varphi$ and $\varphi^{-1}$. For every $m\ge1$, $\opn{Z^{\otimes m}}=\opn{(Z^{\otimes m})^{-1}}=\varphi^{m}$.
\end{lemma}
\begin{proof}
$Z^{\top}Z=\bigl(\begin{smallmatrix}2&1\\1&1\end{smallmatrix}\bigr)$ has characteristic polynomial $\lambda^2-3\lambda+1$, whose roots are $(3\pm\sqrt5)/2=\varphi^{\pm2}$. Since $(A\otimes B)^{\top}(A\otimes B)=A^{\top}A\otimes B^{\top}B$, whose eigenvalues are products of eigenvalues, the spectral norm is multiplicative over Kronecker products. Finally, $Z^{-1}=JZJ$ with the orthogonal matrix $J=\operatorname{diag}(1,-1)$, so $Z^{-1}$ has the same singular values as $Z$, and $(Z^{\otimes m})^{-1}=(Z^{-1})^{\otimes m}$.
\end{proof}

\begin{theorem}[Numerical cut rank]\label{thm:numerical}
For every cut with $L,R\ne\varnothing$ and every $\tau>0$,
\begin{equation}\label{eq:thm2}
\rk{\varphi^{N}\tau}V_{\cut}-2\;\le\;\rk{\tau}C_{\cut}\;\le\;\rk{\varphi^{-N}\tau}V_{\cut}.
\end{equation}
\end{theorem}
\begin{proof}
By~\eqref{eq:unfold} and Lemmas~\ref{lem:sv}(ii,iii) and~\ref{lem:Z}, using $|L|+|R|=N$,
\begin{equation}\label{eq:sandwich}
\varphi^{-N}\sigma_k(D_{\cut})\le\sigma_k(V_{\cut})\le\varphi^{N}\sigma_k(D_{\cut})\qquad(k\ge1).
\end{equation}
\emph{Upper bound.} If $\sigma_k(C)>\tau$, then $\sigma_k(D)\ge\sigma_k(C)>\tau$ by Lemma~\ref{lem:sv}(iv), and so $\sigma_k(V)\ge\varphi^{-N}\sigma_k(D)>\varphi^{-N}\tau$. Singular values are nonincreasing, hence $\rk\tau C\le\rk{\varphi^{-N}\tau}V$. \emph{Lower bound.} Let $r=\rk{\varphi^N\tau}V$. If $r\le2$ there is nothing to prove. Otherwise $\sigma_r(V)>\varphi^N\tau$, so $\sigma_r(D)\ge\varphi^{-N}\sigma_r(V)>\tau$, and Lemma~\ref{lem:sv}(v) with $k=r-2$ gives $\sigma_{r-2}(C)\ge\sigma_r(D)>\tau$. Hence $\rk\tau C\ge r-2$.
\end{proof}

\begin{remark}[Constants]\label{rem:constants}
The factor $\varphi^{N}$ in~\eqref{eq:sandwich} cannot be improved in general. If $D=xy^{\top}$, where $x$ and $y$ are the top right singular vectors of $Z^{\otimes|L|}$ and $Z^{\otimes|R|}$, then $\sigma_1(V)=\varphi^{N}\sigma_1(D)$, which the script confirms to a relative error of $\PhiAtt$ for $N=6$ to $12$. The factors in Theorem~\ref{thm:numerical} are also approached for matrices supported on the cross-cut block. On vectors with vanishing $\varnothing$-entry, the smallest singular value of $Z^{\otimes m}$ is $\varphi^{-m}$ times a factor that decreases from $\RestrMinOne$ at $m=1$ to $\RestrMinFive$ at $m=5$ and $\RestrMinTen$ at $m=10$, and the largest is $\varphi^{m}$ times a factor that increases from $\RestrMaxOne$ to $\RestrMaxFive$ and $\RestrMaxTen$. Over the random games of Section~\ref{sec:numerics}, the largest observed value of $\max_k|\ln(\sigma_k(V)/\sigma_k(D))|/(N\ln\varphi)$ was \ExpMax. The M\"obius basis has condition number $\varphi^{2N}$, so numerical statements in this basis lose resolution exponentially in $N$. For example, at $N=16$ the condition number is $\varphi^{32}\approx4.9\times10^{6}$. This loss is the reason for moving to the Fourier basis in Section~\ref{sec:fourier}.
\end{remark}

For tail-energy rank we use $r_\varepsilon(A;\nu)=\min\{k\ge0:\,\eta_k(A)\le\varepsilon\,\nu^2\}$, which normalizes the tail of $A$ by a reference norm $\nu$.

\begin{proposition}[$\varepsilon$-rank]\label{prop:epsrank}
For every cut with $L,R\ne\varnothing$ and every $\varepsilon>0$,
\[
r_{\varphi^{4N}\varepsilon}\bigl(V_{\cut};\fro{V_{\cut}}\bigr)-2\;\le\;r_{\varepsilon}\bigl(C_{\cut};\fro{D_{\cut}}\bigr)\;\le\;r_{\varphi^{-4N}\varepsilon}\bigl(V_{\cut};\fro{V_{\cut}}\bigr).
\]
\end{proposition}
\begin{proof}
Lemma~\ref{lem:sv}(ii,iii) and Lemma~\ref{lem:Z} give $\varphi^{-2N}\eta_k(D)\le\eta_k(V)\le\varphi^{2N}\eta_k(D)$ for $k\ge0$. The case $k=0$ reads $\varphi^{-2N}\fro{D}^2\le\fro{V}^2\le\varphi^{2N}\fro{D}^2$. \emph{Upper bound.} Let $k=r_{\varphi^{-4N}\varepsilon}(V;\fro V)$. Then $\eta_k(C)\le\eta_k(D)\le\varphi^{2N}\eta_k(V)\le\varphi^{-2N}\varepsilon\fro{V}^2\le\varepsilon\fro{D}^2$. \emph{Lower bound.} Let $k=r_\varepsilon(C;\fro D)$. Lemma~\ref{lem:sv}(v) gives $\eta_{k+2}(D)\le\eta_k(C)\le\varepsilon\fro D^2$, so $\eta_{k+2}(V)\le\varphi^{2N}\varepsilon\fro D^2\le\varphi^{4N}\varepsilon\fro V^2$, that is, $r_{\varphi^{4N}\varepsilon}(V;\fro V)\le k+2$.
\end{proof}

\begin{remark}\label{rem:vacuous}
Since $\varphi^4\approx6.85$, the lower bound in Proposition~\ref{prop:epsrank} is vacuous as soon as $\varphi^{4N}\varepsilon\ge1$, for example for $N\ge4$ when $\varepsilon=10^{-3}$. Only the upper bound carries information at the sizes of interest. Section~\ref{sec:fourier} gives the lossless alternative.
\end{remark}

\section{Fourier basis and interaction indices}\label{sec:fourier}

Let $\Hh=2^{-1/2}\bigl(\begin{smallmatrix}1&1\\1&-1\end{smallmatrix}\bigr)$, which is symmetric, orthogonal and an involution, and set $\F=(\Hh\otimes\cdots\otimes\Hh)\V$. Then
\begin{equation}\label{eq:fourier}
\F_{\mathbf s}=2^{-N/2}\sum_{x}(-1)^{\mathbf s\cdot x}\V_x=2^{N/2}\,\hat v(S),\qquad \hat v(S)=2^{-N}\sum_{X\subseteq[N]}(-1)^{|S\cap X|}v(X),
\end{equation}
with $S=\{i:s_i=1\}$. Here $\hat v$ is the Fourier--Walsh expansion of $v$ viewed as a function on $\{0,1\}^N$, with characters $\omega_S(x)=(-1)^{\sum_{i\in S}x_i}$~\citep{odonnell2014}. The argument of Lemma~\ref{lem:kron} gives $F_{\cut}=\Hh^{\otimes|L|}V_{\cut}(\Hh^{\otimes|R|})^{\top}$, which is an orthogonal equivalence, so $F_{\cut}$ and $V_{\cut}$ have the same singular values. We partition $F_{\cut}$ as in~\eqref{eq:blocks}. Its first row ($S_L=\varnothing$) and first column ($S_R=\varnothing$) form the boundary, and the remaining block is the \emph{cross-cut Fourier matrix} $\Cx_{\cut}=\bigl(\F_{S_L\cup S_R}\bigr)_{S_L\ne\varnothing,S_R\ne\varnothing}$.

\begin{theorem}[Fourier version]\label{thm:fourier}
For every cut with $L,R\ne\varnothing$, every $\tau>0$ and every $\varepsilon>0$,
\begin{align*}
&\rank\Cx_{\cut}\le\chi_{\cut}\le\rank\Cx_{\cut}+2,\\
&\rk\tau V_{\cut}-2\le\rk\tau\Cx_{\cut}\le\rk\tau V_{\cut},\\
&r_\varepsilon\bigl(V_{\cut};\fro{V_{\cut}}\bigr)-2\le r_\varepsilon\bigl(\Cx_{\cut};\fro{V_{\cut}}\bigr)\le r_\varepsilon\bigl(V_{\cut};\fro{V_{\cut}}\bigr).
\end{align*}
\end{theorem}
\begin{proof}
$\sigma_k(F_{\cut})=\sigma_k(V_{\cut})$ for all $k$, and $\fro{F_{\cut}}=\fro{V_{\cut}}$. Apply Lemma~\ref{lem:sv}(iv,v) to the pair $(\Cx,F)$ exactly as in the proofs of Theorems~\ref{thm:exact} and~\ref{thm:numerical} and Proposition~\ref{prop:epsrank}, with all conditioning factors equal to one.
\end{proof}

The Banzhaf interaction index of a coalition $T$ is~\citep{grabisch1999}
\begin{equation}\label{eq:banzhaf}
\IB(T)=2^{-(N-|T|)}\sum_{S\subseteq[N]\setminus T}\Delta_Tv(S),\qquad \Delta_Tv(S)=\sum_{W\subseteq T}(-1)^{|T|-|W|}v(S\cup W).
\end{equation}
It is positive for complementary (synergistic) coalitions. For example $\IB(T)=1$ for the unanimity game $u_T$.

\begin{lemma}[Fourier--Banzhaf identity]\label{lem:banzhaf}
For every $T\subseteq[N]$,
\[
\IB(T)=(-2)^{|T|}\,\hat v(T)=(-2)^{|T|}\,2^{-N/2}\,\F_T .
\]
\end{lemma}
\begin{proof}
Every $X\subseteq[N]$ can be written uniquely as $X=S\cup W$ with $S=X\setminus T$ and $W=X\cap T$. Hence
$\sum_{S\subseteq[N]\setminus T}\Delta_Tv(S)=\sum_{X}(-1)^{|T|-|X\cap T|}v(X)=(-1)^{|T|}\,2^N\hat v(T)$ by~\eqref{eq:fourier}. Multiplying by $2^{-(N-|T|)}$ gives $(-1)^{|T|}2^{|T|}\hat v(T)$.
\end{proof}

The sign factor $(-1)^{|T|}$ arises because the character takes the value $-1$ when a player is present, $x_i=(1-\omega_{\{i\}}(x))/2$. The factor $2^{|T|}$ compensates for the averaging over the $2^{N-|T|}$ contexts in~\eqref{eq:banzhaf}. The identity agrees with the dividend formula $\IB(T)=\sum_{U\supseteq T}2^{-(|U|-|T|)}d_U$~\citep{grabisch2000}. Both were checked numerically against the definition~\eqref{eq:banzhaf} (Section~\ref{sec:numerics}).

\begin{corollary}[Cross-cut Banzhaf matrix]\label{cor:banzhaf}
Let $B^\times_{\cut}=\bigl(\IB(T_L\cup T_R)\bigr)_{T_L\ne\varnothing,T_R\ne\varnothing}$. Then $B^\times_{\cut}=2^{-N/2}P_L\Cx_{\cut}P_R$ with $P_L=\operatorname{diag}\bigl((-2)^{|T_L|}\bigr)$ and $P_R=\operatorname{diag}\bigl((-2)^{|T_R|}\bigr)$. Consequently $\rank B^\times_{\cut}=\rank\Cx_{\cut}$, and
\[
\rank B^\times_{\cut}\le\chi_{\cut}\le\rank B^\times_{\cut}+2,\qquad 2^{2-N/2}\sigma_k(\Cx_{\cut})\le\sigma_k(B^\times_{\cut})\le2^{N/2}\sigma_k(\Cx_{\cut}).
\]
\end{corollary}
\begin{proof}
The factorization follows from Lemma~\ref{lem:banzhaf}, since $(-2)^{|T_L\cup T_R|}=(-2)^{|T_L|}(-2)^{|T_R|}$. The diagonal factors are invertible, which gives equality of ranks. The entries of $P_L$ have absolute values between $2$ and $2^{|L|}$, and similarly for $P_R$, so Lemma~\ref{lem:sv}(ii,iii) gives the singular-value bounds.
\end{proof}

\begin{figure}[tbp]
\centering
\includegraphics[width=\linewidth]{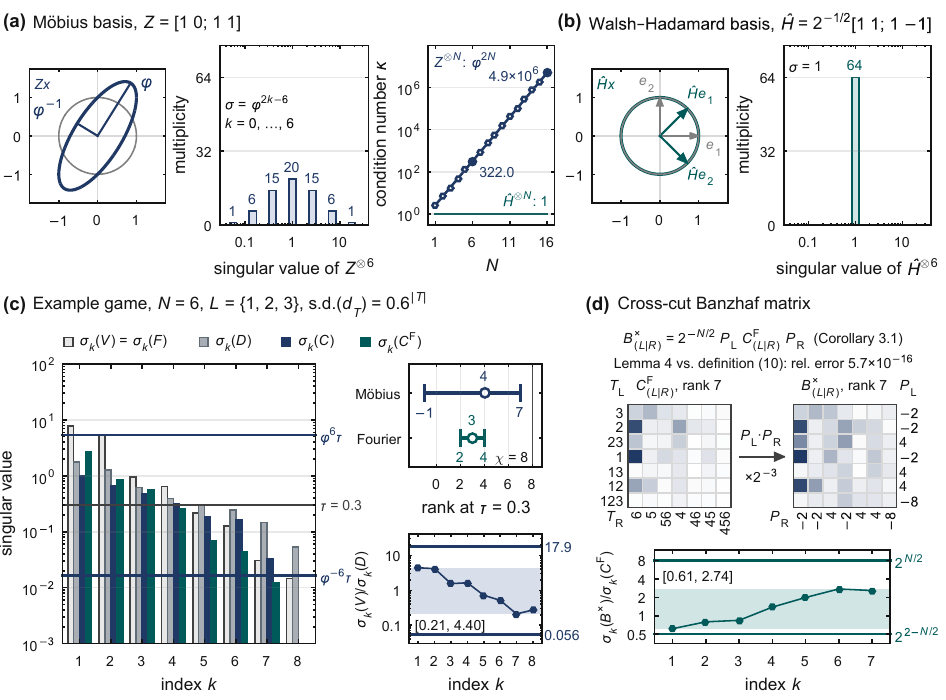}
\caption{M\"obius basis and orthonormal Walsh--Hadamard (Fourier) basis. (a)~$Z$ maps the unit circle (grey) onto an ellipse with semi-axes $\varphi$ and $\varphi^{-1}$. The singular values of $Z^{\otimes6}$ are $\varphi^{2k-6}$ with multiplicity $\binom{6}{k}$ (bar labels). Right: the condition number of $Z^{\otimes N}$ is $\varphi^{2N}$ (Lemma~\ref{lem:Z}), that is $\varphi^{12}\approx322.0$ at $N=6$ and $\varphi^{32}\approx4.9\times10^{6}$ at $N=16$, whereas that of $\Hh^{\otimes N}$ is one. This conditioning produces the factors $\varphi^{\pm N}$ of Theorem~\ref{thm:numerical}. (b)~$\Hh$ is orthogonal: it maps the unit circle onto itself and sends $e_1,e_2$ to $\Hh e_1,\Hh e_2$, so all $64$ singular values of $\Hh^{\otimes6}$ equal one and Theorem~\ref{thm:fourier} holds without loss. (c)~Singular values of the unfoldings for one real-valued game with $N=6$, $L=\{1,2,3\}$ and independent Gaussian dividends $d_T$ with standard deviation $0.6^{|T|}$ (fixed random seed). At $\tau=0.3$ (thin dark line) the M\"obius statement~\eqref{eq:thm2}, with thresholds $\varphi^{\pm6}\tau$ on $V_{\cut}$ (thick navy lines), reads $-1\le\rk{\tau}C_{\cut}=4\le7$. The Fourier statement of Theorem~\ref{thm:fourier}, with the same threshold $\tau$ on both sides, reads $2\le\rk{\tau}\Cx_{\cut}=3\le4$ (intervals at the upper right; $\chi_{\cut}=8$). Lower right: the ratios $\sigma_k(V_{\cut})/\sigma_k(D_{\cut})$ lie in $[0.21,4.40]$ (shaded), inside the interval $[\varphi^{-6},\varphi^{6}]=[0.056,17.9]$ allowed by~\eqref{eq:sandwich}. (d)~Corollary~\ref{cor:banzhaf} for the same game. The cross-cut Banzhaf matrix $B^\times_{\cut}$ is the cross-cut Fourier matrix $\Cx_{\cut}$ multiplied by $2^{-N/2}$ and rescaled by $P_L=P_R=\operatorname{diag}\bigl((-2)^{|T|}\bigr)$; rows and columns of $\Cx_{\cut}$ are labelled by $T_L$ and $T_R$, those of $B^\times_{\cut}$ by the diagonal entries of $P_L$ and $P_R$, and shading is proportional to the absolute value of each entry. Both matrices have rank $7$, the ratios $\sigma_k(B^\times_{\cut})/\sigma_k(\Cx_{\cut})$ lie in $[0.61,2.74]\subset[2^{2-N/2},2^{N/2}]=[0.5,8]$, and Lemma~\ref{lem:banzhaf} matches the brute-force definition~\eqref{eq:banzhaf} to a relative error of $5.7\times10^{-16}$.}
\label{fig:bases}
\end{figure}

Exact cut rank thus counts, to within two, both the linearly independent cross-cut Harsanyi patterns and the linearly independent cross-cut Banzhaf patterns. In particular $|\rank C_{\cut}-\rank B^\times_{\cut}|\le2$. Numerically, the Fourier statement is lossless, but it refers to Banzhaf interactions. The statement about Harsanyi dividends pays the factor $\varphi^{\pm N}$. Figure~\ref{fig:bases} compares the two bases on a worked example.

The Banzhaf transform is local because its weight $2^{-(|U|-|T|)}$ factorizes over the players in $U\setminus T$: $\IB=K_{1/2}^{\otimes N}\D$ with $K_x=\bigl(\begin{smallmatrix}1&x\\0&1\end{smallmatrix}\bigr)$. The Shapley interaction index~\citep{grabisch1999} has the dividend form $\ISh(T)=\sum_{U\supseteq T}d_U/(|U|-|T|+1)$~\citep{grabisch2000}, whose weight does not factorize. It is an average of local transforms instead, in the same way as Owen's diagonal formula averages the Shapley value over the multilinear extension~\citep{owen1972}.

\begin{proposition}[Shapley interaction index]\label{prop:sii}
Let $\mathcal I$ be the tensor with entries $\ISh(T)$ and $I_{\cut}$ its unfolding over a cut with $L,R\ne\varnothing$.
\begin{enumerate}[label=(\roman*),leftmargin=*,itemsep=1pt]
\item $\mathcal I=\int_0^1K_x^{\otimes N}\D\,dx$, and $\rank I_{\cut}\le\bigl(\min(|L|,|R|)+1\bigr)\chi_{\cut}\le\lceil(N+1)/2\rceil\,\chi_{\cut}$. The first bound is attained by $u_{[N]}$.
\item For the unanimity game $u_T$ with $|T\cap L|=a\ge0$ and $|T\cap R|=b\ge0$, $\rank I_{\cut}=\min(a,b)+1$, whereas $\chi_{\cut}=1$.
\item $\chi_{\cut}\le\bigl(\min(|L|,|R|)+1\bigr)\rank I_{\cut}$, with equality at every cut for the game with $\ISh(U)=\one[U=[N]]$.
\end{enumerate}
Hence neither $\rank I_{\cut}\le\chi_{\cut}+c$ nor $\chi_{\cut}\le\rank I_{\cut}+c$ holds with $c$ independent of $N$, in contrast with Corollary~\ref{cor:banzhaf}.
\end{proposition}
The proof is in Appendix~\ref{app:sii}. On \SiiN{} random games with $N=5$ to $8$ (with the definition of \citet{grabisch1999} evaluated by brute force on every third of them), the dividend formula matched the definition to a relative error of $\SiiErr$, the bound (i) held in every case, and $\rank I_{\cut}-\chi_{\cut}$ reached \SiiExcess. Item (ii) was confirmed for all \SiiUnaN{} pairs $a,b\le6$ with $T=[N]$. Item (iii) held on the same random games (\SiiConvViol{} violations), where the inverse transform of its proof reproduced the dividends to a relative error of $\SiiInvErr$, and its equality case was confirmed at balanced cuts for $N=2$ to $12$.

\section{R\'enyi entropies of the value tensor}\label{sec:entropy}

Entropies require a choice of gauge. Adding a constant to a game, $v\mapsto v+c$, changes only $d_\varnothing$ (by $c$) and $\F_\varnothing$ (by $2^{N/2}c$). The matrices $C_{\cut}$ and $\Cx_{\cut}$ do not change, but the spectrum of $V_{\cut}$ does. Adding a large constant to any game drives its entropy towards zero, because the constant contributes a dominant rank-one term. Entropies are therefore defined only after fixing this freedom. We use the mean-zeroed tensor $\bar\V=\V-2^{-N}\sum_x\V_x$, which is equivalent to $\F_\varnothing=0$. Theorems~\ref{thm:exact}--\ref{thm:fourier} apply to $\bar\V$ unchanged, because its $C$ and $\Cx$ are those of $\V$. For $\bar\V\ne0$ let $\bar F_{\cut}$ be the Fourier unfolding of $\bar\V$ and define
\begin{equation}\label{eq:renyi}
\lambda_k=\frac{\sigma_k(\bar V_{\cut})}{\fro{\bar V_{\cut}}},\qquad S^{(\alpha)}_{\cut}=\frac{1}{1-\alpha}\ln\sum_k\lambda_k^{2\alpha},\qquad \gamma_{\cut}=\frac{\fro{\Cx_{\cut}}^2}{\fro{\bar F_{\cut}}^2},
\end{equation}
for $\alpha\in(0,1)\cup(1,\infty)$, with the limits $S^{(1)}=-\sum_k\lambda_k^2\ln\lambda_k^2$ (von Neumann entropy) and $S^{(\infty)}=-\ln\lambda_1^2$. $S^{(\alpha)}$ is nonincreasing in $\alpha$. By Parseval, $\fro{\bar F_{\cut}}^2=2^N\operatorname{Var}_S[v(S)]=2^N\sum_{T\ne\varnothing}\hat v(T)^2$, where $S$ is a uniformly random coalition. Hence $\gamma_{\cut}=\sum_{T\cap L\ne\varnothing\ne T\cap R}\hat v(T)^2\big/\sum_{T\ne\varnothing}\hat v(T)^2$ is the share of the variance of $v$ carried by Walsh components that straddle the cut. By Lemma~\ref{lem:banzhaf} it is also a weighted share of squared Banzhaf interactions.

\begin{theorem}[R\'enyi-2 upper bound]\label{thm:renyi}
Let $\bar\V\ne0$ and $L,R\ne\varnothing$. Then $\lambda_1^2+\lambda_2^2\ge1-\gamma_{\cut}$, and, if $\gamma_{\cut}<1$,
\begin{equation}\label{eq:thm4}
S^{(2)}_{\cut}\;\le\;\ln2-2\ln\bigl(1-\gamma_{\cut}\bigr).
\end{equation}
\end{theorem}
\begin{proof}
Let $E$ be $\bar F_{\cut}$ with the block $\Cx_{\cut}$ set to zero. Then $E=e_1f_R^{\top}+f_Le_1^{\top}$ has rank at most two, and $E$ and $\bar F_{\cut}-E$ have disjoint supports. By Lemma~\ref{lem:sv}(i) and orthogonal invariance, $\eta_2(\bar V_{\cut})=\eta_2(\bar F_{\cut})\le\fro{\bar F_{\cut}-E}^2=\fro{\Cx_{\cut}}^2$. Dividing by $\fro{\bar F_{\cut}}^2$ gives $\sum_{k>2}\lambda_k^2\le\gamma$, that is, $\lambda_1^2+\lambda_2^2\ge1-\gamma$. Therefore $\sum_k\lambda_k^4\ge\lambda_1^4+\lambda_2^4\ge\tfrac12(\lambda_1^2+\lambda_2^2)^2\ge\tfrac12(1-\gamma)^2$, and $S^{(2)}=-\ln\sum_k\lambda_k^4$ gives~\eqref{eq:thm4}.
\end{proof}

The right-hand side of~\eqref{eq:thm4} grows without limit as $\gamma_{\cut}\to1$, whereas $S^{(\alpha)}_{\cut}\le\ln\rank\bar V_{\cut}\le\min(|L|,|R|)\ln2$ holds for every game and every $\alpha$. For $\gamma_{\cut}$ close to one the dimension bound is the binding one, and Theorem~\ref{thm:renyiall} below combines the two through $\gamma_n=\min(\gamma,1-2/n)$.

\begin{remark}[Sharpness]\label{rem:sharp}
(a) \emph{At $\gamma=0$.} Let $v(S)=\one[i\in S]+\one[j\in S]$ with $i\in L$ and $j\in R$. Then $\bar V_{\cut}=\alpha\one^{\top}+\one\beta^{\top}$ with $\alpha\perp\one$, $\beta\perp\one$ and $\fro{\alpha\one^{\top}}=\fro{\one\beta^{\top}}$. The two singular values are equal and $\gamma=0$, so $S^{(\alpha)}=\ln2$ for every $\alpha$, which is the value of the bound.
(b) \emph{Asymptotically, for every $\gamma\in(0,1)$.} Let $n=2^{\min(|L|,|R|)}\ge4$. Choose orthonormal $u_0=e_\varnothing,u_1,\dots,u_{n-1}$ in the row space and $w_0=e_\varnothing,w_1,\dots,w_{n-1}$ in the column space, and set
$\bar F=\sqrt{(1-\gamma)/2}\,(u_1w_0^{\top}+u_0w_1^{\top})+\sqrt{\gamma/(n-2)}\sum_{j=2}^{n-1}u_jw_j^{\top}$.
This expression is already a singular value decomposition. Its first two terms lie in the boundary, the corner entry vanishes, and the remaining terms lie in the cross-cut block. Every real matrix is the Fourier unfolding of some game, because $\Hh$ is an involution. Hence $\gamma_{\cut}=\gamma$ and $S^{(2)}=-\ln\bigl((1-\gamma)^2/2+\gamma^2/(n-2)\bigr)$, which is below~\eqref{eq:thm4} by $\ln\bigl(1+2\gamma^2/((n-2)(1-\gamma)^2)\bigr)\to0$ as $n\to\infty$. Section~\ref{sec:numerics} confirms this construction numerically for $N=12$.
\end{remark}

The bound~\eqref{eq:thm4} extends to every R\'enyi order, and the extension shows a sharp difference between $\alpha>1$ and $\alpha\le1$. For an integer $n\ge2$ and $\gamma\in[0,1]$ put $\gamma_n=\min(\gamma,1-2/n)$ and
\begin{equation}\label{eq:Phi}
\Phi_\alpha(\gamma;n)=\frac{1}{1-\alpha}\ln\Bigl(2^{1-\alpha}(1-\gamma_n)^\alpha+(n-2)^{1-\alpha}\gamma_n^\alpha\Bigr)\qquad(\alpha\in(0,1)\cup(1,\infty)),
\end{equation}
with $\Phi_1(\gamma;n)=(1-\gamma_n)\ln2+h(\gamma_n)+\gamma_n\ln(n-2)$, where $h(g)=-g\ln g-(1-g)\ln(1-g)$, and $\Phi_\infty(\gamma;n)=\ln2-\ln(1-\gamma_n)$. Terms with $n-2=0$ are read as zero.

\begin{theorem}[R\'enyi entropies at a given cross-cut share]\label{thm:renyiall}
Let $\bar\V\ne0$, $L,R\ne\varnothing$, and let $n\ge2$ be an integer with $n\ge\rank\bar V_{\cut}$, for example $n=2^{\min(|L|,|R|)}$.
\begin{enumerate}[label=(\roman*),leftmargin=*,itemsep=1pt]
\item $S^{(\alpha)}_{\cut}\le\Phi_\alpha(\gamma_{\cut};n)$ for every $\alpha\in(0,\infty]$.
\item For $\alpha\in(1,\infty]$ and $\gamma_{\cut}<1$, $\Phi_\alpha(\gamma_{\cut};n)\le\ln2-\frac{\alpha}{\alpha-1}\ln(1-\gamma_{\cut})$, with $\frac{\alpha}{\alpha-1}=1$ for $\alpha=\infty$. The case $\alpha=2$ is Theorem~\ref{thm:renyi}, and $\alpha=\infty$ gives $S^{(\infty)}\le\ln2-\ln(1-\gamma_{\cut})$.
\item If $n=2^{\min(|L|,|R|)}\ge4$ and $0\le\gamma\le1-2/n$, there is a game with $\gamma_{\cut}=\gamma$ and $S^{(\alpha)}_{\cut}=\Phi_\alpha(\gamma;n)$ for every $\alpha\in(0,\infty]$.
\end{enumerate}
\end{theorem}
The proof is in Appendix~\ref{app:renyi}. Part (iii) shows that $\Phi_\alpha(\gamma;2^{\min(|L|,|R|)})$ is the exact maximum of $S^{(\alpha)}$ over all games on a cut of the given size with $\gamma_{\cut}=\gamma\le1-2^{1-\min(|L|,|R|)}$. Since $S^{(\alpha)}$ is nonincreasing in $\alpha$, Theorem~\ref{thm:renyi} already implies $S^{(\alpha)}\le\ln2-2\ln(1-\gamma)$ for $\alpha\ge2$. Part (ii) improves this to $\ln2-\frac{\alpha}{\alpha-1}\ln(1-\gamma)$, which is smaller for $\alpha>2$, and extends it to $1<\alpha<2$. The value at $\gamma=0$ is $\ln2$ for every $\alpha$, because the unfolding then has rank at most two.

\begin{corollary}[No bound in terms of $\gamma$ for $\alpha\le1$]\label{cor:alphasmall}
Let $\alpha\in(0,1]$ and $\delta\in(0,1)$. For every $\ell\ge2$ with $\delta\le1-2^{1-\ell}$ there is a game with $N=2\ell$, $|L|=|R|=\ell$ and $\gamma_{\cut}=\delta$ such that
\[
S^{(\alpha)}_{\cut}\;\ge\;\ln\bigl(2^\ell-2\bigr)+\frac{\alpha}{1-\alpha}\ln\delta\quad(\alpha<1),\qquad S^{(1)}_{\cut}\;\ge\;\delta\ln\frac{2^\ell-2}{\delta}.
\]
In particular $\sup\{S^{(\alpha)}_{\cut}:\gamma_{\cut}\le\delta\}=\infty$ for every $\delta>0$, so no function of $\gamma$ alone bounds $S^{(\alpha)}$ when $\alpha\le1$. For $\alpha>1$, in contrast, $S^{(\alpha)}\le\ln2-\frac{\alpha}{\alpha-1}\ln(1-\delta)$ whenever $\gamma_{\cut}\le\delta$, for every $N$.
\end{corollary}
\begin{proof}
Take the game of Theorem~\ref{thm:renyiall}(iii) with $n=2^\ell$ and $\gamma=\delta$. For $\alpha<1$, dropping the first term inside the logarithm in~\eqref{eq:Phi} gives $\Phi_\alpha\ge\frac{1}{1-\alpha}\ln\bigl((n-2)^{1-\alpha}\delta^\alpha\bigr)$. For $\alpha=1$, $h(\delta)\ge-\delta\ln\delta$ gives $\Phi_1\ge\delta\ln((n-2)/\delta)$. Both lower bounds tend to infinity as $\ell\to\infty$. The last claim is Theorem~\ref{thm:renyiall}(i,ii), since $\ln2-\frac{\alpha}{\alpha-1}\ln(1-\gamma)$ is increasing in $\gamma$.
\end{proof}

Table~\ref{tab:growth} shows the dichotomy on the games of Theorem~\ref{thm:renyiall}(iii) with $\gamma=0.3$, computed through the full pipeline from game to tensor, centring, transform and singular value decomposition. Every entry equals $\Phi_\alpha(0.3;2^k)$ to $10^{-9}$ (\FiveAttChecks{} checks over $\gamma\in\{0.05,0.3,0.6\}$, \FiveAttViol{} failures). For $\alpha>1$ the columns approach the $N$-independent bounds of Theorem~\ref{thm:renyiall}(ii), while for $\alpha\le1$ they keep growing with $k$.

\begin{table}[t]
\centering\small
\caption{Maximal R\'enyi entropies at $\gamma=0.3$ (Theorem~\ref{thm:renyiall}(iii)), $N=2k$, $|L|=|R|=k$, $n=2^k$. The $N$-independent bounds of Theorem~\ref{thm:renyiall}(ii) are $\GrowthLimOneHalf$, $\GrowthLimTwo$, $\GrowthLimThree$ and $\GrowthLimInf$ for $\alpha=3/2,2,3,\infty$; there is no such bound for $\alpha=1/2$ and $\alpha=1$. Entropies in nats.}
\label{tab:growth}
\begin{tabular}{cccccccc}
\toprule
$k$ & $n$ & $S^{(1/2)}$ & $S^{(1)}$ & $S^{(3/2)}$ & $S^{(2)}$ & $S^{(3)}$ & $S^{(\infty)}$\\
\midrule
\TableGrowth
\bottomrule
\end{tabular}
\end{table}

The next result shows that $\gamma$ controls the entropy from above only. Interaction mass across a cut, however large, does not force entanglement.

\begin{proposition}[No lower bound in terms of $\gamma$]\label{prop:nolower}
For every $\delta>0$ there is a game and a cut with $\gamma_{\cut}\ge1-\delta$ and $S^{(2)}_{\cut}\le\delta$. Consequently, if $f:[0,1)\to\R$ is nondecreasing and $S^{(2)}_{\cut}\ge f(\gamma_{\cut})$ for all mean-zeroed games and cuts, then $f\le0$ on $[0,1)$. In particular no lower bound $S^{(2)}\ge f(\gamma)$ with $f$ nondecreasing and unbounded exists.
\end{proposition}
\begin{proof}
Take the unanimity game $u_T$ with $|T\cap L|=a\ge1$ and $|T\cap R|=b\ge1$, where $N\ge a+b$ and the remaining players are arbitrary. Since $x_i=(1-\omega_{\{i\}}(x))/2$,
\[
u_T(x)=\prod_{i\in T}\frac{1-\omega_{\{i\}}(x)}{2}=2^{-(a+b)}\sum_{U\subseteq T}(-1)^{|U|}\omega_U(x),
\]
so $\hat u_T(U)=(-1)^{|U|}2^{-(a+b)}$ for $U\subseteq T$ and $\hat u_T(U)=0$ otherwise. All nonzero coefficients have the same magnitude. After mean-zeroing, which removes $U=\varnothing$, $2^{a+b}-1$ equal-magnitude coefficients remain, and $(2^a-1)(2^b-1)$ of them straddle the cut. Therefore
\begin{equation}\label{eq:gamma-unanimity}
\gamma_{\cut}(u_T)=\frac{(2^a-1)(2^b-1)}{2^{a+b}-1}.
\end{equation}
For the entropy, let $\mu=2^{-(a+b)}$ be the mean of $u_T$. Then $\bar U=h_Lh_R^{\top}-\mu\one\one^{\top}$, where $h_L$ and $h_R$ are the indicator vectors of $\{x_L:T\cap L\subseteq X_L\}$ and $\{x_R:T\cap R\subseteq X_R\}$, so $\rank\bar U\le2$. We have $\one^{\top}h_L=\|h_L\|^2$ and $\one^{\top}h_R=\|h_R\|^2$, which gives $\langle\bar U,h_Lh_R^{\top}\rangle=\|h_L\|^2\|h_R\|^2(1-\mu)$, while $\fro{\bar U}^2=\|h_L\|^2\|h_R\|^2-2^N\mu^2=\|h_L\|^2\|h_R\|^2(1-\mu)$. Since $\sigma_1(\bar U)\ge\langle\bar U,h_Lh_R^{\top}\rangle/(\|h_L\|\|h_R\|)$,
\[
\lambda_1^2\ge\frac{\|h_L\|^2\|h_R\|^2(1-\mu)^2}{\fro{\bar U}^2}=1-\mu,\qquad\text{hence}\qquad S^{(2)}_{\cut}\le-\ln\lambda_1^4\le-2\ln\bigl(1-2^{-(a+b)}\bigr).
\]
With $a=b=\ell$, \eqref{eq:gamma-unanimity} gives $\gamma=(2^\ell-1)/(2^\ell+1)\to1$ while $S^{(2)}\le-2\ln(1-4^{-\ell})\to0$. This proves the first claim. For the second, fix $\gamma_0<1$ and $\delta>0$, and pick a member of the family with $\gamma\ge\gamma_0$ and $S^{(2)}\le\delta$. Then $f(\gamma_0)\le f(\gamma)\le S^{(2)}\le\delta$.
\end{proof}

\begin{figure}[tbp]
\centering
\includegraphics[width=\linewidth]{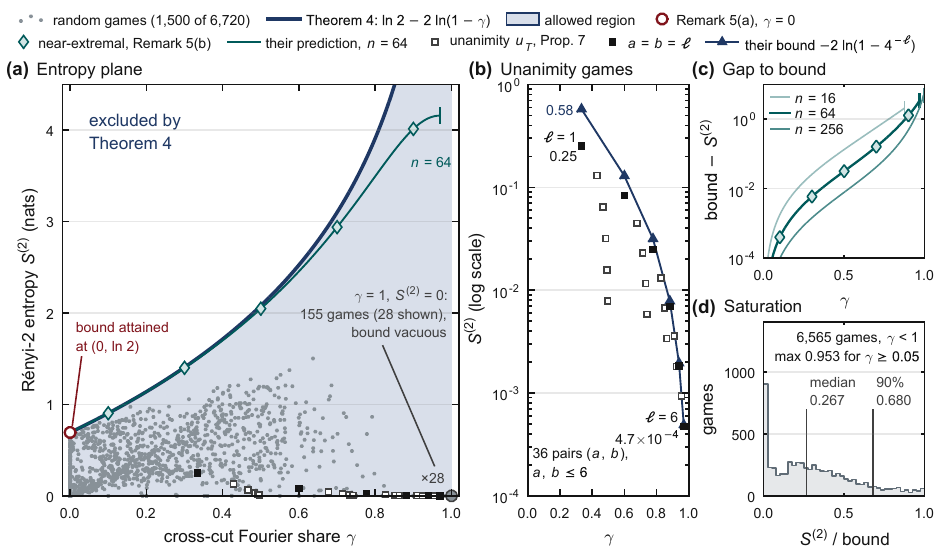}
\caption{R\'enyi-2 entropy $S^{(2)}_{\cut}$ of the mean-zeroed value tensor against the cross-cut Fourier share $\gamma_{\cut}$. (a)~The thick curve is the bound $\ln2-2\ln(1-\gamma)$ of Theorem~\ref{thm:renyi}, and the shaded area is the region that it allows. Dots are a stratified random subsample of $1{,}500$ of the \NInst{} random games of Section~\ref{sec:numerics}, $125$ from each of the twelve families, with the same seed as Tables~\ref{tab:thm1} and~\ref{tab:thm2}. The $155$ multiplicative games with $\gamma=1$, for which the bound is vacuous, have $S^{(2)}=0$ and lie in the lower right corner ($28$ of them are in the subsample). Diamonds are the near-extremal games of Remark~\ref{rem:sharp}(b) with $N=12$, $|L|=6$ and $n=64$ (Table~\ref{tab:tight}). The thin curve is their predicted entropy $-\ln\bigl((1-\gamma)^2/2+\gamma^2/62\bigr)$, drawn up to $\gamma=1-2/n$, beyond which the construction is no longer extremal (Theorem~\ref{thm:renyiall}(iii)). The open circle at $(0,\ln2)$ is the example of Remark~\ref{rem:sharp}(a), at which the bound is attained. Squares are the unanimity games $u_T$ of Proposition~\ref{prop:nolower} for all \PFiveN{} pairs $(a,b)$ with $a,b\le6$ and $a+b\le12$; filled squares are the diagonal $a=b=\ell$. (b)~The same unanimity games on a logarithmic entropy scale. The line with triangles is the bound $-2\ln(1-4^{-\ell})$ of the diagonal, which equals $0.58$ at $\ell=1$. Along the diagonal $\gamma$ rises from $0.33$ to $0.97$ while $S^{(2)}$ falls from $0.25$ to $4.7\times10^{-4}$, so a large cross-cut share does not force entanglement. (c)~Distance $\ln\bigl(1+2\gamma^2/((n-2)(1-\gamma)^2)\bigr)$ between the bound and the near-extremal construction for $n=16$, $64$ and $256$. Each curve ends at $\gamma=1-2/n$, and the gap shrinks as $n$ grows. The diamonds are the measured gaps of the five games in (a), from $4.0\times10^{-4}$ at $\gamma=0.1$ to $1.28$ at $\gamma=0.9$. (d)~Ratio $S^{(2)}/(\ln2-2\ln(1-\gamma))$ for the $6{,}565$ random games with $\gamma<1$. The median is \FourMed{} and the 90th percentile is \FourNinety{}. The maximum $1$ occurs at $\gamma=0$; for $\gamma\ge0.05$ it is \MaxRatioPos{}. All points are computed by verify.py.}
\label{fig:entropy}
\end{figure}

A single high-order interaction is one cross-cut pattern. It has rank one in $C_{\cut}$ and carries a large Fourier mass, yet it produces almost no entanglement. Rank and entropy count independent patterns, whereas $\gamma$ measures how much interaction mass crosses the cut. Two-sided entropy bounds therefore have to depend on the spectrum of $\Cx_{\cut}$, not only on its norm (Problem~\ref{prob:spectral}). Figure~\ref{fig:entropy} places the bound, the near-extremal games, the unanimity family and the random games of Section~\ref{sec:numerics} in the $(\gamma,S^{(2)})$ plane.

\section{Numerical verification}\label{sec:numerics}

The script verify.py (\RepoURL, folder reproduce; requires only NumPy and SciPy) tests the theorems and propositions above, the items of Corollary~\ref{cor:exact} that apply to its random families, and both the rank part and the singular-value part of Corollary~\ref{cor:banzhaf}. It uses no model and no data. Twelve random families were generated, each with $80$ instances for every $N\in\{6,\dots,12\}$, giving \NInst{} games in total. The integer-valued families are as follows. dense\_int has i.i.d.\ dividends in $\{-3,\dots,3\}$. sparse\_int has $s\in\{1,\dots,2N\}$ nonzero dividends at random coalitions. pairwise\_int has dividends only for $|T|\le2$. separable\_int has $C=0$. lowrank\_int has a cross-cut block equal to a sum of $0$ to $5$ integer outer products and a random boundary. multiplicative\_int uses $a_i\in\{-2,-1,0,2,3\}$. unanimity contains unanimity games $u_T$ with random $T$. The real-valued families are as follows. dense\_gauss has Gaussian dividends. decay has Gaussian dividends scaled by $\rho^{|T|}$ with $\rho\in\{0.2,0.35,0.5\}$. lowrank\_noisy is the real-valued analogue of lowrank\_int, perturbed by Gaussian noise of size $10^{-u}$ with $u\sim U(2,8)$. multiplicative\_float uses $a_i\sim U(0.2,3)$ kept away from $1$. sparse\_float has Gaussian dividends whose magnitudes span four decades. For each game the cut size $|L|$ was drawn uniformly from $\{1,\dots,N-1\}$. Half of the cuts are contiguous and half are random subsets. The structured checks of Sections~\ref{sec:exact} and~\ref{sec:classes} use a second random stream, so they do not change the random families.

\subsection{Exact ranks}
Theorem~\ref{thm:exact} holds over every field (Remark~\ref{rem:field}). For the integer families we computed $\rank C$ and $\chi=\rank V$ exactly over $\mathrm{GF}(p)$ with $p=10^9+7$. This is an exact test of~\eqref{eq:thm1} over that field. Floating-point ranks used the tolerance $4\max(m,n)\,\epsilon_{\text{mach}}\,\opn{\cdot}$, measured against the norm of the full unfolding. They agreed with the $\mathrm{GF}(p)$ ranks in every integer instance. The results are in Table~\ref{tab:thm1}. The column ``cor.'' checks item (a) on separable\_int, both bounds of item (b) with $k=2$ on pairwise\_int, both parts of item (c) on the sparse families, item (d) on unanimity, and $\chi=2$, $\rank C=1$ and the dividend count of item (e) on the multiplicative families.

\begin{table}[t]
\centering\small
\caption{Theorem~\ref{thm:exact} and Corollary~\ref{cor:exact} on \NInst{} games ($N=6$ to $12$). ``viol.''\ counts violations of~\eqref{eq:thm1}. The three middle columns give the distribution of $\chi-\rank C$. ``cor.''\ counts violations of the applicable item of Corollary~\ref{cor:exact} out of the number of applicable games. The last column shows agreement of floating-point and $\mathrm{GF}(p)$ ranks.}
\label{tab:thm1}
\begin{tabular}{lrrrrrcc}
\toprule
family & games & viol. & $\chi-\rank C=0$ & $=1$ & $=2$ & cor. & $\mathrm{GF}(p)=$ float\\
\midrule
\TableOne
\bottomrule
\end{tabular}
\end{table}

\subsection{Numerical ranks}
For Theorem~\ref{thm:numerical}, $12$ thresholds were drawn per game, log-uniformly in the interval $[10^{-10}\sigma_1,\,2\varphi^N\sigma_1]$, with $\sigma_1=\max(\sigma_1(C),\sigma_1(V))$. Up to six singular values of $C$ above the floor were added as boundary cases. Thresholds below $10^{-10}\sigma_1$ were not tested, because floating-point singular value decompositions resolve singular values only to about $\epsilon_{\text{mach}}\sigma_1$. Proposition~\ref{prop:epsrank} was tested at $\varepsilon\in\{10^{-1},10^{-2},10^{-3},10^{-6}\}$, and Theorem~\ref{thm:fourier} in the same way without conditioning factors. Table~\ref{tab:thm2} reports the counts. The intermediate inequalities were also checked directly. There were \SvSandwichViol{} violations of the sandwich~\eqref{eq:sandwich} and \InterViol{} violations of the interlacing inequalities of Lemma~\ref{lem:sv}(iv,v). The singular values of $F_{\cut}$ and $V_{\cut}$ agreed to a relative error of $\SvErrF$. On \BzN{} games with $N=6$ to $8$, Lemma~\ref{lem:banzhaf} matched the brute-force definition~\eqref{eq:banzhaf} to a relative error of $\BzErr$, and the dividend formula to $\BzErrDiv$. The ranks of $B^\times_{\cut}$ and $\Cx_{\cut}$ differed in \BzRank{} cases, and the singular-value bounds of Corollary~\ref{cor:banzhaf} failed in \BzSvViol{} cases.

\subsection{Tolerances}
Comparisons used a relative slack of $10^{-9}$, and $10^{-7}$ for the Fourier thresholds. The option -{}-strict sets every slack to zero. With it, no check of a theorem or proposition fails. The only failures are \ExcSandwich{} checks of the sandwich~\eqref{eq:sandwich}, \ExcInterlace{} interlacing checks and \ExcLambda{} checks of $\lambda_1^2+\lambda_2^2\ge1-\gamma$, which are intermediate inequalities, and \ExcBanzhaf{} checks of the singular-value bounds of Corollary~\ref{cor:banzhaf}. All of them occur at cases of equality, and the largest excess, relative to the largest singular value of the matrix involved, is $\ExcMax$, which is rounding error.

\begin{table}[t]
\centering\small
\caption{Numerical-rank statements. A check is one (game, threshold) pair. The last two columns give the fraction of checks in which the lower or upper bound holds with equality. They are not given for Proposition~\ref{prop:epsrank}, whose lower bound is non-vacuous in only \EpsNonVac{} of its checks (Remark~\ref{rem:vacuous}).}
\label{tab:thm2}
\begin{tabular}{lrrcc}
\toprule
statement & checks & violations & eq.\ lower & eq.\ upper\\
\midrule
\TableTwo
\bottomrule
\end{tabular}
\end{table}

\subsection{Entropy}
Theorem~\ref{thm:renyi} was evaluated on all \FourN{} mean-zeroed games. There were \FourViol{} violations of~\eqref{eq:thm4} and \FourLamViol{} violations of $\lambda_1^2+\lambda_2^2\ge1-\gamma$. The ratio $S^{(2)}/(\ln2-2\ln(1-\gamma))$ had median \FourMed{}, 90th percentile \FourNinety{} and maximum \FourMax. Among the \FourGzeroN{} games with $\gamma=0$ the largest entropy was \FourGzeroMax{}, which equals $\ln2$ to the printed precision. Among random games with $\gamma\ge0.05$ the largest ratio was \MaxRatioPos. The example of Remark~\ref{rem:sharp}(a) gives $S^{(2)}=\TightSTwo$. Table~\ref{tab:tight} reproduces the construction of Remark~\ref{rem:sharp}(b) through the full pipeline. Theorem~\ref{thm:renyiall}(i) was checked on the same games for $\alpha\in\{1/2,1,3/2,2,3,\infty\}$ with $n=2^{\min(|L|,|R|)}$ (\FiveChecks{} checks, \FiveViol{} violations), and the bound of part (ii) in \FiveInfChecks{} checks with $\alpha>1$ and $\gamma<1$ (\FiveInfViol{} violations). Table~\ref{tab:p5} confirms~\eqref{eq:gamma-unanimity} and the entropy bound of Proposition~\ref{prop:nolower} for all \PFiveN{} pairs $(a,b)$ with $a,b\le6$ and $a+b\le12$, at $N=\PFiveNmin$ to $\PFiveNmax$ (\PFiveViol{} violations). It also confirms the equal-magnitude Fourier support.

\subsection{Structured classes}
Table~\ref{tab:structured} summarizes the checks of Sections~\ref{sec:exact}--\ref{sec:fourier}. For Proposition~\ref{prop:support} every triple $(|L|,|R|,k)$ with $2\le k\le N\le12$ was tested twice: with Gaussian dividends of degree at most $k$, for which $\chi=\kappa_k$ and $\rank C=\rho_k$ must hold exactly, and with integer dividends in $\{-3,\dots,3\}$ over $\mathrm{GF}(p)$, for which the bounds must hold. The integer games attained both values in \DegIntEq{} of the \DegCases{} cases. The chain of inequalities~\eqref{eq:grouped} was checked for all $N\le30$. In total the run reported \TotalViol{} violations and took about \RunTime{}\,s on a desktop computer.

\begin{table}[t]
\centering\small
\caption{Near-extremal games of Remark~\ref{rem:sharp}(b) ($N=12$, $|L|=6$, $n=64$). ``predicted'' is \mbox{$-\ln\bigl((1-\gamma)^2/2+\gamma^2/(n-2)\bigr)$}. Entropies in nats.}
\label{tab:tight}
\begin{tabular}{ccccc}
\toprule
target $\gamma$ & measured $\gamma$ & $S^{(2)}$ & predicted & bound~\eqref{eq:thm4}\\
\midrule
\TableTight
\bottomrule
\end{tabular}
\end{table}

\begin{table}[t]
\centering\small
\caption{Unanimity games $u_T$ with $|T\cap L|=a$, $|T\cap R|=b$ (diagonal $a=b$ shown). ``formula'' is~\eqref{eq:gamma-unanimity} and ``$S^{(2)}$ bound'' is $-2\ln(1-2^{-(a+b)})$. Entropies in nats.}
\label{tab:p5}
\begin{tabular}{cccccccc}
\toprule
$a$ & $b$ & $N$ & $\gamma$ & formula & $S^{(2)}$ & $S^{(2)}$ bound & rank\\
\midrule
\TableFive
\bottomrule
\end{tabular}
\end{table}

\begin{table}[t]
\centering\small
\caption{Checks of the structured results. Integer-valued games use exact ranks over $\mathrm{GF}(p)$.}
\label{tab:structured}
\begin{tabular}{llrr}
\toprule
statement & instances & checks & violations\\
\midrule
Prop.~\ref{prop:support}(ii,iii), generic & all $(|L|,|R|,k)$, $N\le12$ & \DegCases & \DegGenViol\\
Prop.~\ref{prop:support}(i,iii), integer & all $(|L|,|R|,k)$, $N\le12$ & \DegCases & \DegIntViol\\
inequalities~\eqref{eq:grouped} & all $(|L|,|R|,k)$, $N\le30$ & \DegCasesThirty & \DegOrderViol\\
Prop.~\ref{prop:blocksym}, symmetric & random $g$ and cut, $N=6$--$12$ & \SymN & \SymViol\\
Prop.~\ref{prop:blocksym}(a), generic full rank & Gaussian $g$ & \SymN & \SymGenMiss\\
Prop.~\ref{prop:blocksym}(b), threshold & all $q$, all contiguous cuts & \ThrN & \ThrViol\\
Prop.~\ref{prop:blocksym}(c), exactly-$w$ & all $w$, all contiguous cuts & \DickeN & \DickeViol\\
Prop.~\ref{prop:blocksym}(d), glove & $|L|,|R|\le6$ & \GloveN & \GloveViol\\
Prop.~\ref{prop:addpair}(a), additive & random weights and zero patterns & \AddN & \AddViol\\
Prop.~\ref{prop:addpair}(b), pairwise & five graph types, \PwDP{} graph games & \PwN & \PwViol\\
Prop.~\ref{prop:addpair}(b), path, $\chi\le3$ & contiguous cuts & \PwChainN & \PwChainViol\\
Prop.~\ref{prop:addpair}(b), $\rank W_{LR}=\nu(G_{LR})$ & Gaussian weights & \PwGenN & \PwGenViol\\
Prop.~\ref{prop:ttparams}, Tables~\ref{tab:dictionary},~\ref{tab:profiles} & eight families, $N=8$ & $8$ & \ProfViol\\
Prop.~\ref{prop:sii}(i) & random games, $N=5$--$8$ & \SiiN & \SiiViol\\
Prop.~\ref{prop:sii}(ii) & $u_{[N]}$, $a,b\le6$ & \SiiUnaN & \SiiUnaViol\\
Prop.~\ref{prop:sii}(iii) & random games, $N=5$--$8$ & \SiiN & \SiiConvViol\\
Prop.~\ref{prop:sii}(iii), equality & balanced cuts, $N=2$--$12$ & \SiiSharpN & \SiiSharpViol\\
Thm.~\ref{thm:renyiall}(i) & all random games, six orders & \FiveChecks & \FiveViol\\
Thm.~\ref{thm:renyiall}(iii), attainment & $k=2$--$6$, three values of $\gamma$ & \FiveAttChecks & \FiveAttViol\\
\bottomrule
\end{tabular}
\end{table}

\FloatBarrier
\section{A pilot measurement on six language models}\label{sec:pilot}

To see where real value functions fall in the picture of Sections~\ref{sec:exact}--\ref{sec:entropy}, we measured exact value tensors of \PilotModelsWord{} instruction-tuned LLMs from three families: Qwen3 with $1.7$, $4$ and $8$ billion parameters~\citep{qwen3}, OLMo-2-Instruct with $1$ and $7$ billion~\citep{olmo2}, and Mistral-7B-Instruct-v0.3~\citep{mistral7b}. This is a pilot, meant to show what the quantities of this paper look like on real models. It is not a systematic study.

\subsection{Setup}
The players are the $N=12$ word tokens of a sentence from the Stanford Sentiment Treebank (SST-2)~\citep{socher2013}. Punctuation is not a player and is always kept. The \PilotSent{} sentences are the first \PilotNeg{} sentences with exactly twelve words that contain a negator (\emph{not}, \emph{n't}, \emph{no}, \emph{never}, \emph{nothing} and similar) and the first \PilotNeg{} that contain none, taken in the order of the validation split and then of the unlabelled test split. For a coalition $S$ the model receives an instruction to classify the sentiment of the review as positive or negative, with the words outside $S$ deleted, and
\[
v(S)=\ln\sum_{a\in\mathcal A_+}p(a\mid S)-\ln\sum_{a\in\mathcal A_-}p(a\mid S),
\]
where $\mathcal A_\pm$ are the first tokens of the answers ``positive'' and ``negative'' in their capitalized and uncapitalized forms, with and without a leading space (for Mistral-7B-Instruct the capitalized answers begin with the tokens ``Pos'' and ``Neg''). All $2^{12}=4{,}096$ coalitions were evaluated, so every tensor is exact up to floating-point error. The models ran in half precision on one graphics card, with the thinking mode of Qwen3 disabled, and took between about two and ten minutes per model. To estimate the floating-point error, $256$ random coalitions per sentence were evaluated a second time in a different batch composition. For every sentence the mean squared difference, relative to the variance of $v$, was at most $\PilotEtaMax$, at least five times smaller than the tolerances $\varepsilon\ge10^{-3}$ used below. On the \PilotLab{} labelled sentences the sign of $v([N])$ agrees with the label in \PilotAccRange{} cases, depending on the model (Table~\ref{tab:pilot}). All quantities refer to the mean-zeroed tensor $\bar\V$ and the cuts $\{1,\dots,k\}|\{k+1,\dots,N\}$ of the word order. The $\varepsilon$-rank $r_\varepsilon$ of an unfolding is the smallest $r$ with tail energy $\eta_r\le\varepsilon\fro{\bar V}^2$.

\begin{figure}[tbp]
\centering
\includegraphics[width=\linewidth]{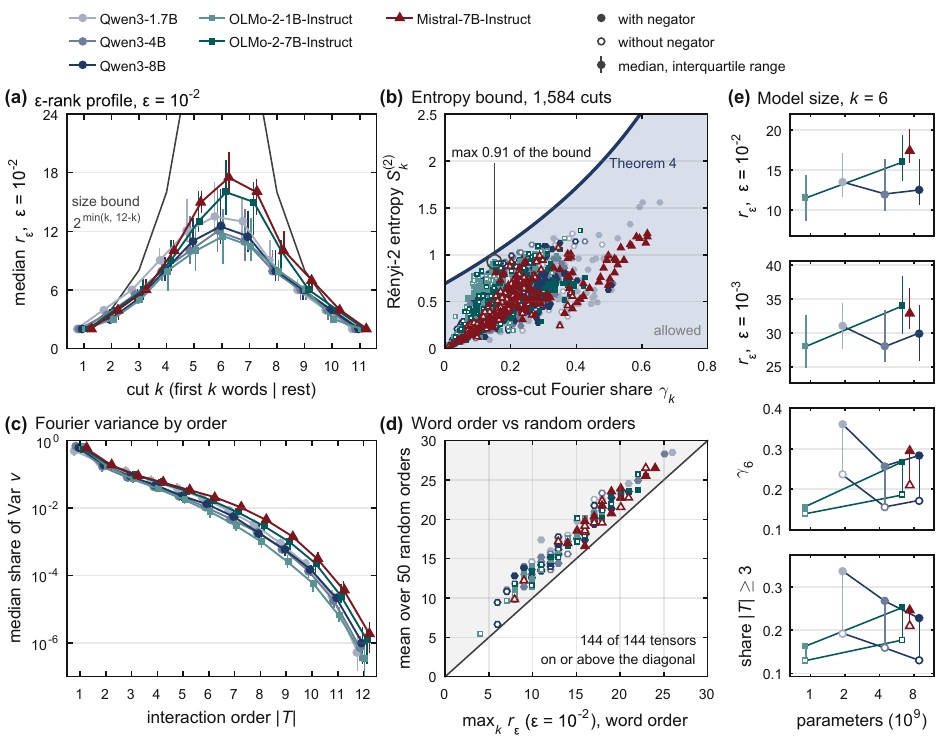}
\caption{Value tensors of \PilotModelsWord{} instruction-tuned LLMs on \PilotSent{} SST-2 sentences with $N=12$ word players, computed from all $4{,}096$ coalitions. Colour marks the family (Qwen3 navy circles, OLMo 2 teal squares, Mistral crimson triangles), and lighter shades mark smaller models of a family. Filled markers are sentences with a negator, open markers sentences without. (a)~Median $\varepsilon$-rank of the mean-zeroed unfolding at each cut of the word order for $\varepsilon=10^{-2}$, with interquartile bars. The thin line is the size bound $2^{\min(k,12-k)}$, clipped at the top of the axis. (b)~Cross-cut Fourier share $\gamma_k$ and R\'enyi-2 entropy $S^{(2)}_k$ for all $11$ cuts of every sentence and model (\PilotCuts{} cuts). The thick curve is the bound $\ln2-2\ln(1-\gamma)$ of Theorem~\ref{thm:renyi}, and the shaded area is the region that it allows. The ringed cut comes closest to the bound, at \PilotRatMax{} of it. (c)~Median share of the variance of $v$ carried by Fourier coefficients of each interaction order $|T|$, with interquartile bars. (d)~Largest $\varepsilon$-rank ($\varepsilon=10^{-2}$) over the cuts of the word order against its mean over $50$ uniformly random orders of the players. Points on or above the diagonal (shaded), here all \PilotTensors{} tensors, mean that the word order needs no larger bonds than a random order. (e)~Quantities at the middle cut $k=6$ against model size: median $r_{10^{-2}}$ and $r_{10^{-3}}$ with interquartile bars, and the medians of $\gamma_6$ and of the variance share of orders $|T|\ge3$ over sentences with (filled) and without (open) a negator. Models of the same size are offset slightly along the size axis.}
\label{fig:llm}
\end{figure}

\begin{table}[t]
\centering\small
\caption{Summary of the pilot (Figure~\ref{fig:llm}). ``acc.'' is the number of labelled sentences whose sign of $v([N])$ matches the label, $e_{\mathrm{fp}}$ the largest relative floating-point error, $r_{10^{-2}}$ and $r_{10^{-3}}$ the median $\varepsilon$-ranks at the middle cut $k=6$, ``max'' the largest $r_{10^{-2}}$ over all cuts and sentences, $\gamma_6$ and $S^{(2)}_6$ the medians at the middle cut, ``ratio'' the largest value of $S^{(2)}_k$ divided by the bound of Theorem~\ref{thm:renyi}, $|T|\ge3$ the median variance share of interaction orders three and higher, ``neg.\,/\,other'' the median $\gamma_6$ over sentences with and without a negator, and the last column the median over sentences of the largest $r_{10^{-2}}$ over the cuts of the word order and of its mean over $50$ random orders.}
\label{tab:pilot}
\footnotesize\setlength{\tabcolsep}{3.0pt}
\begin{tabular}{lccccccccccc}
\toprule
model & acc. & $e_{\mathrm{fp}}$ & $r_{10^{-2}}$ & $r_{10^{-3}}$ & max & $\gamma_6$ & $S^{(2)}_6$ & ratio & $|T|\ge3$ & neg.\,/\,other & word / random\\
\midrule
\TablePilot
\bottomrule
\end{tabular}
\end{table}

\subsection{Findings}
The checks of Theorems~\ref{thm:fourier}, \ref{thm:renyi} and~\ref{thm:renyiall} hold on all \PilotCuts{} cuts, with \PilotViol{} violations. Being theorems, they must hold, so this only confirms that the pipeline computes the quantities of this paper correctly. Six observations go beyond that.

First, the value tensors are not of low rank at fine tolerances, but they are far from generic. In every model the $\varepsilon$-rank rises towards the middle of the sentence and falls again towards its end (Figure~\ref{fig:llm}(a)). At the middle cut its median is \PilotRTwoRange{} at $\varepsilon=10^{-2}$ and \PilotRThreeRange{} at $\varepsilon=10^{-3}$ across the models, out of $64$, and no unfolding of any sentence has $r_{10^{-2}}$ above \PilotRMaxAll.

Second, most of the variance is carried by low interaction orders, but not all of it. Across the models, single words carry a median share of \PilotDegOneRange{}, pairs \PilotDegTwoRange{}, and orders three and higher \PilotDegThreeRange{}, with the shares falling by a factor of about three from single words to pairs and of about two per order thereafter (Figure~\ref{fig:llm}(c)). A model that keeps only interactions of order at most two would therefore miss a share of \PilotDegThreeRange{} of the variance, although the cut ranks stay moderate. This is the situation of Section~\ref{sec:sparsity}, in which low rank and sparsity are different structures.

Third, Theorem~\ref{thm:renyi} is informative on these tensors without being tight. The median cross-cut share at the middle cut is \PilotGamRange{} and the median R\'enyi-2 entropy there \PilotSRange{} nats. Over all \PilotCuts{} cuts, $S^{(2)}_k$ never exceeds \PilotRatMax{} of the bound, and the median ratio per model is \PilotRatMedRange{} (Figure~\ref{fig:llm}(b)).

Fourth, sentences with a negator tend to have more interaction across the middle of the sentence, but the evidence is weak. In all \PilotModels{} models their median $\gamma_6$ is larger than that of the sentences without a negator, and in all of them their share of interaction orders three and higher is larger as well (Table~\ref{tab:pilot}). Taken one model at a time, the difference is significant at the $5\%$ level in \PilotGamSig{} models for $\gamma_6$ and in \PilotDegSig{} for the high-order share (two-sided Mann--Whitney tests with \PilotNeg{} sentences per group, $p$ from \PilotGamPRange{} and from \PilotDegPRange{}, not corrected for multiple comparisons; after a Holm correction over the \PilotModels{} models the number of significant differences drops to \PilotGamHolm{} for $\gamma_6$ and \PilotDegHolm{} for the high-order share). The differences in $S^{(2)}_6$ are not significant in any model (smallest $p=\PilotSPMin$). All models see the same \PilotSent{} sentences, so the comparisons are not independent, and the agreement in direction is an exploratory observation, not a tested hypothesis.

Fifth, the word order is a better order for a TT than a random one. For each of the \PilotTensors{} tensors the largest $\varepsilon$-rank over the cuts of the word order is at most its mean over $50$ random orders (Figure~\ref{fig:llm}(d)), and the medians per model are \PilotNatRange{} against \PilotRandRange. This is what Section~\ref{sec:order} predicts if interacting words tend to be close in the sentence, so that contiguous cuts separate fewer of them; the pilot does not test this explanation.

Finally, within the range from $1$ to $8$ billion parameters none of these quantities changes consistently with model size (Figure~\ref{fig:llm}(e)). The three Qwen3 models differ little, while OLMo-2-7B-Instruct has larger middle-cut ranks and shares than OLMo-2-1B-Instruct.

The pilot covers \PilotModelsWord{} models from three families with $1$ to $8$ billion parameters, one task, one prompt, deletion as the way of removing words, and \PilotSent{} short sentences. It does not show how the ranks scale with $N$, which is the question of Problem~\ref{prob:llm}. The scripts and the measured tensors are available in the same repository (\RepoURL).

\section{Open problems}\label{sec:open}

The following problems are stated for a game $v$ on $N$ players, a cut $L|R$ with $L,R\ne\varnothing$, and the quantities of~\eqref{eq:renyi}. Write $\mu^\times$ for the vector of squared singular values of $\Cx_{\cut}$ divided by $\fro{\Cx_{\cut}}^2$, and $S^{(\alpha)}(\Cx)$ for its R\'enyi entropy.

\begin{problem}[Spectral two-sided entropy bounds]\label{prob:spectral}
By Lemma~\ref{lem:sv}(iv,v) the spectra of $\bar F_{\cut}$ and $\Cx_{\cut}$ interlace with shift two. For $\alpha\in(0,1)$ this already gives a two-sided bound: summing $\sigma_k(\Cx_{\cut})^{2\alpha}\le\sigma_k(\bar F_{\cut})^{2\alpha}$ and $\sigma_{k+2}(\bar F_{\cut})^{2\alpha}\le\sigma_k(\Cx_{\cut})^{2\alpha}$ yields $\gamma_{\cut}^\alpha\le\sum_k\lambda_k^{2\alpha}\big/\sum_k(\mu^\times_k)^\alpha\le\gamma_{\cut}^\alpha+2$, hence $\bigl|S^{(\alpha)}_{\cut}-S^{(\alpha)}(\Cx)\bigr|\le\max\bigl(\alpha\ln(1/\gamma_{\cut}),\ln3\bigr)/(1-\alpha)$. For $\alpha\ge1$ no bound of this form holds: the games of Remark~\ref{rem:sharp}(b) have $S^{(\alpha)}(\Cx)=\ln(n-2)$, whereas $S^{(\alpha)}_{\cut}$ stays bounded for $\alpha>1$ by Theorem~\ref{thm:renyiall}(ii) and equals $\gamma\ln(n-2)+O(1)$ for $\alpha=1$. Does
\[
0\;\le\;S^{(1)}_{\cut}-\gamma_{\cut}\,S^{(1)}(\Cx)\;\le\;\ln3
\]
hold for every game and cut, and which functional of the spectrum of $\Cx_{\cut}$ gives two-sided bounds on $S^{(\alpha)}_{\cut}$ for $\alpha>1$?
Unanimity games (Proposition~\ref{prop:nolower}) show that $\gamma$ alone does not suffice for a lower bound, and Theorem~\ref{thm:renyiall} shows that it does not suffice for an upper bound when $\alpha\le1$.
\end{problem}

\begin{problem}[R\'enyi orders below one]\label{prob:alphasmall}
For $\alpha<1$ neither $\gamma$ nor a tail condition on $\Cx_{\cut}$ bounds the entropy. The game of Corollary~\ref{cor:alphasmall} with $|L|=|R|=\ell$ and share $\delta=\min(\gamma,\varepsilon)$ has $\gamma_{\cut}\le\gamma$ and $\eta_r(\Cx_{\cut})\le\eta_0(\Cx_{\cut})=\delta\fro{\bar F_{\cut}}^2\le\varepsilon\fro{\bar F_{\cut}}^2$ for every $r\ge0$, while $S^{(\alpha)}_{\cut}\ge\ln(2^\ell-2)+\frac{\alpha}{1-\alpha}\ln\delta$ is unbounded in $\ell$. Orders $\alpha<1$ control MPS truncation errors~\citep{schuch2008}, so the question becomes structural. For games of degree at most $k\ge2$ on a cut with $|L|=a$ and $|R|=b$, determine
\[
\Phi^{(k)}_\alpha(\gamma;a,b)=\sup\Bigl\{S^{(\alpha)}_{\cut}:\ d_T=0\text{ for }|T|>k,\ \gamma_{\cut}=\gamma\Bigr\},
\]
which lies between $0$ and $\ln\kappa_k(a,b)$, and decide whether $\Phi^{(k)}_\alpha(\gamma;a,b)-\ln\rho_k(a,b)$ is bounded uniformly in $a$ and $b$ for fixed $\alpha$, $\gamma$ and $k$.
\end{problem}

\begin{problem}[Shapley interaction blocks]\label{prob:sii}
Proposition~\ref{prop:sii} shows that the exact ranks of $I_{\cut}$ and $V_{\cut}$ determine each other up to the factor $\min(|L|,|R|)+1$ in each direction, and that both factors are attained. Determine the numerical counterpart: the best constants in singular-value sandwiches between $I_{\cut}$, its cross-cut block $(\ISh(T_L\cup T_R))_{T_L,T_R\ne\varnothing}$ and $V_{\cut}$, which involve the conditioning of the linear map from dividends to $\ISh$ and of its inverse. Are they polynomial or exponential in $N$?
\end{problem}

\begin{problem}[Query complexity of cross interpolation]\label{prob:tci}
Cut rank alone cannot bound the number of value queries. The game $v(S)=\one[S=S_0]$ has $\chi=1$ at every cut, yet a deterministic algorithm that only queries values needs $2^N-1$ queries in the worst case over $S_0$ to locate the nonzero entry, because an adversary can answer $0$ while two or more candidates remain. Let $c_k$ be the cross-cut ranks of Proposition~\ref{prop:ttparams}. Find a condition on the dividends, for example a bound on $\max_x|\V_x|/(2^{-N/2}\fro{\V})$, under which an adaptive cross-interpolation algorithm recovers $\V$ exactly with high probability from
\[
N^{O(1)}\max_k(c_k+2)^2
\]
queries, and express the condition in terms of the dividends or interaction indices.
\end{problem}

\begin{problem}[Generic rank beyond fixed supports]\label{prob:generic}
Proposition~\ref{prop:support} gives the generic cut rank when the support of the dividends is prescribed and the nonzero dividends are free. Many games of interest have constrained nonzero dividends, for example $d_T=\delta(|T|)$ for symmetric games or $d_T=\prod_{i\in T}(a_i-1)$ for multiplicative ones. For a family of games parametrized polynomially by $\theta\in\R^p$, characterize the generic value of $\chi_{\cut}(v_\theta)$ in terms of the parametrization, and decide whether it can be computed in time polynomial in $N$ and $p$.
\end{problem}

\begin{problem}[Rank profiles of LLM value tensors]\label{prob:llm}
For a prompt with $N$ words let $v(S)$ be the log-odds of the model's prediction when only the words in $S$ are kept. Measure, along the word order, the profiles $k\mapsto\rk{\tau\fro{\bar F}}\Cx_k$, $\gamma_k$ and $S^{(\alpha)}_k$ for $\alpha\in\{1/2,1,2\}$ and a fixed relative threshold $\tau$. For families of prompts of growing length, decide whether
\[
\max_k\rk{\tau\fro{\bar F}}\Cx_k\ \text{ is bounded, polynomial or exponential in }N,
\]
which is the analogue of an area law~\citep{eisert2010}, and whether syntactic constructions such as negation scope or long-distance agreement change the scaling. The pilot of Section~\ref{sec:pilot} gives these profiles for $N=12$ and weak evidence that negation raises the cross-cut share.
\end{problem}

\section{Discussion}\label{sec:discussion}

\subsection{Bond dimension and query cost}
For the natural player order, the smallest TT rank at the cut $\{1,\dots,k\}\,|\,\{k+1,\dots,N\}$ equals $\chi$ at that cut~\citep{oseledets2011}. Theorem~\ref{thm:exact} therefore fixes the entire TT rank profile of a game, to within two at each bond, by the ranks of its cross-cut dividend matrices, and Proposition~\ref{prop:ttparams} turns this into two-sided bounds on the size of the smallest exact TT. Applying $Z^{-1}$ or $\Hh$ to each TT core converts a TT of $v$ into a TT of the dividends or of $\hat v$ with identical ranks. In tree- and graph-structured networks the product of the bond dimensions on every edge separator is bounded below by the same rank~\citep{heidari2026tnshapg}, and Theorem~\ref{thm:exact} ties that bound to $\rank C_{(A|B)}$ to within two. Cross interpolation with exact pivots represents a tensor of TT ranks $r_k$ through $\sum_k2r_{k-1}r_k$ of its entries, and practical TCI uses on the order of $N\chi^2$ queries per sweep~\citep{oseledets2010,savostyanov2014,nunez2025}. The number of independent cross-cut interaction patterns thus sets the size of a TT surrogate of the value function, although finding the pivots can require many more queries (Problem~\ref{prob:tci}). Theorems~\ref{thm:numerical} and~\ref{thm:fourier} explain how truncation tolerances carry over. A tolerance on values says little about dividend ranks at the same tolerance when $N$ is large. It transfers exactly to the cross-cut Fourier block, and to the cross-cut Banzhaf block up to the factors $2^{2-N/2}$ and $2^{N/2}$ of Corollary~\ref{cor:banzhaf}.

\subsection{Player order and tree-structured networks}\label{sec:order}
A TT fixes a linear order of the players, and its rank profile depends on that order. Theorem~\ref{thm:exact} holds for every bipartition, so for each order it bounds the TT ranks by the cross-cut dividend matrices of the contiguous cuts of that order. The pairwise game on the path $1-2-\dots-N$ with nonzero weights shows how large the effect can be. In the natural order every contiguous cut is crossed by one edge and $\chi\le3$ (Proposition~\ref{prop:addpair}(b)). If the odd players are placed before the even ones, the middle cut separates the odd from the even players, every edge crosses it, and $W_{LR}$ is a bidiagonal matrix with nonzero diagonal, so $\rank W_{LR}=\lfloor N/2\rfloor$ and $\lfloor N/2\rfloor\le\chi\le\lfloor N/2\rfloor+2$, against $\chi\le3$ in the natural order. Every edge of a tree tensor network also defines a bipartition, so Theorem~\ref{thm:exact} bounds its bond dimensions in the same way~\citep{heidari2026tnshapg}. Choosing an order or a tree is therefore choosing which cuts must carry the interactions, and a good choice keeps the strongly interacting players on the same side of as many cuts as possible. For text, the surface order of the words is the natural choice, but long-distance syntactic dependencies cross many contiguous cuts, and a tree that follows the dependency structure may give smaller bond dimensions. Whether it does is part of Problem~\ref{prob:llm}.

\subsection{Sparsity and low rank are different structures}\label{sec:sparsity}
By Corollary~\ref{cor:exact}(c), a game with $s$ nonzero dividends has $\chi\le s$ at every cut, so sparsity implies low cut rank. The converse fails. Multiplicative games, in which $\log(v+1)$ is additive, have $2^N-1$ nonzero dividends and $\chi=2$ everywhere. Threshold, majority and glove games are dense with $\chi\le\min(|L|,|R|)+1$ (Proposition~\ref{prop:blocksym}), and games whose dividends are dense but of degree at most two have $\chi\le2+\min(|L|,|R|)$. Methods that assume sparsity of the M\"obius or Fourier spectrum~\citep{kang2024,kang2025,butler2025,gorji2025} and methods that assume low TT rank~\citep{ballester2022,heidari2026tnshap,heidari2026tnshapg} therefore address different classes of games. The class $\{\chi\le s\}$ strictly contains the $s$-sparse games.

\subsection{A measurement programme}
For an LLM, $v(S)$ can be taken as the log-odds of the model's predicted label when only the words in $S$ are kept. For $N\le16$ this requires $2^N$ forward passes and gives exact value tensors. On such tensors, $\rk\tau\Cx$, the cut rank, $\gamma$ and $S^{(\alpha)}$ can be compared across cuts and across linguistic constructions (Problem~\ref{prob:llm}). The results of this paper fix how these quantities relate to one another. Corollary~\ref{cor:banzhaf} says what thresholded Fourier ranks mean in terms of interactions. Theorems~\ref{thm:renyi} and~\ref{thm:renyiall} and Proposition~\ref{prop:nolower} say that a high cross-cut interaction share is necessary but not sufficient for high R\'enyi entropy of order $\alpha>1$, and that for $\alpha\le1$ no function of the share alone bounds the entropy (Corollary~\ref{cor:alphasmall}). R\'enyi-2 entropy can also be estimated by randomized measurements when the tensor is loaded as a quantum state~\citep{brydges2019}. Section~\ref{sec:pilot} carries out a first, small version of this programme.

\subsection{Limitations}
The exact results are elementary linear algebra. Their value lies in identifying the right blocks, in making the constants explicit and in the consequences for structured classes. The M\"obius-basis numerical statements lose a factor $\varphi^{\pm N}$, which is inherent in the conditioning of $Z^{\otimes N}$ (Remark~\ref{rem:constants}). The lower bound of Proposition~\ref{prop:epsrank} is vacuous at practical sizes (Remark~\ref{rem:vacuous}). The entropy results are upper bounds in terms of $\gamma$ and the cut size. They are exact maxima, but they do not use the spectrum of $\Cx_{\cut}$, and two-sided spectral bounds remain open for $\alpha\ge1$ (Problem~\ref{prob:spectral}); for $\alpha<1$ they follow from interlacing. The ranks of the value and Shapley-interaction unfoldings bound each other only up to a factor $\min(|L|,|R|)+1$, which is attained in both directions, and the numerical counterpart is open (Proposition~\ref{prop:sii}, Problem~\ref{prob:sii}). The verification of Section~\ref{sec:numerics} uses only synthetic games, and its floating-point checks depend on the stated tolerances. The exact checks are carried out over $\mathrm{GF}(p)$, where they test Theorem~\ref{thm:exact} exactly because the theorem holds over every field. Reduction modulo $p$ cannot increase the rank of an integer matrix, so these ranks are lower bounds on the ranks over $\mathbb{Q}$, and the floating-point ranks agreed with them in every integer instance.

\section*{Code and data availability}
The verification script, the scripts and measured value tensors of the language-model pilot, and CutRank, a Python library that implements the computations of this paper (cut ranks, cross-cut dividend and Fourier blocks, R\'enyi entropies and their bounds, tensor trains and a search for player orders with small tensor trains), are available at \RepoURL{} under the MIT license.

\bibliographystyle{unsrtnat}
\bibliography{refs}

\appendix
\section{Proofs}\label{app:proofs}

\subsection{Proposition~\ref{prop:support}(iii)}\label{app:support}

\begin{alemma}[Nested neighbourhoods]\label{lem:nested}
Let $G$ be a bipartite graph with row vertices $y_1,\dots,y_m$ whose neighbourhoods are nested, $N(y_1)\subseteq N(y_2)\subseteq\cdots\subseteq N(y_m)$, and let $n_p=|N(y_p)|$ and $n_0=0$. Then
\[
\nu(G)=\min_{0\le p\le m}\bigl(n_p+m-p\bigr).
\]
\end{alemma}
\begin{proof}
\emph{Upper bound.} For each $p$, the columns in $N(y_p)$ together with the rows $y_{p+1},\dots,y_m$ cover every edge, because the rows $y_1,\dots,y_p$ have neighbourhoods inside $N(y_p)$. A matching has at most one edge at each covering vertex, so $\nu(G)\le n_p+m-p$. \emph{Lower bound.} Process the rows in the order $y_1,\dots,y_m$ and match $y_p$ to an unused column of $N(y_p)$ if one exists. Let $M_p$ be the number of matched rows among $y_1,\dots,y_p$, with $M_0=0$. All columns used so far lie in $N(y_{p-1})\subseteq N(y_p)$. If $n_p>M_{p-1}$, an unused column exists and $M_p=M_{p-1}+1$. Otherwise $n_p\le M_{p-1}$, and since $M_{p-1}\le n_{p-1}\le n_p$ we have $M_p=M_{p-1}=n_p$. In both cases $M_p=\min(M_{p-1}+1,n_p)$, and by induction $M_p=\min_{0\le q\le p}(n_q+p-q)$. The greedy procedure produces a matching of size $M_m$, which equals the upper bound.
\end{proof}

\begin{proof}[Proof of Proposition~\ref{prop:support}(iii)]
Let $a=|L|$, $b=|R|$ and $\mathcal T=\{T:|T|\le k\}$. In $G^\times_{\mathcal T}$ the row $T_L$ with $|T_L|=j$ is adjacent to the columns $T_R\ne\varnothing$ with $|T_R|\le k-j$, so the neighbourhoods are nested and depend only on $j$. Order the rows by decreasing $j$. Within a block of equal $j$ the expression $n_p+m-p$ of Lemma~\ref{lem:nested} decreases in $p$, so the minimum is attained at $p=0$ or at the end of a block. If the last block processed is $\{|T_L|\ge t\}$ with $1\le t\le k$, then $n_p=\sum_{i=1}^{k-t}\binom bi$ and $m-p=\sum_{j=1}^{t-1}\binom aj$. Rows with $j\ge k$ have empty neighbourhoods, and the choice $p=0$ gives the total number of rows, which is not smaller than the value at $t=k$. Hence $\nu(G^\times_{\mathcal T})=\rho_k(a,b)$. For $G_{\mathcal T}$ the same argument with $T_L=\varnothing$ and $T_R=\varnothing$ allowed, and $0\le t\le k+1$, gives $\kappa_k(a,b)$. For~\eqref{eq:grouped}, a matching uses at most $\min\bigl(\binom aj,\sum_{i=1}^{k-j}\binom bi\bigr)$ edges at the rows with $|T_L|=j$, which gives the first inequality and, by symmetry, its counterpart with $L$ and $R$ exchanged. The second inequality drops the inner minimum.
\end{proof}

\subsection{Proposition~\ref{prop:blocksym}(a)--(d)}\label{app:blocksym}

Let $a=|L|$ and $b=|R|$. Each item computes $\rank H$ for the matrix $H$ of the main statement.

(a) For $v(S)=g(|S|)$ we have $h(s,t)=g(s+t)$ at every cut, so $H$ is a Hankel matrix. Let $m=\min(a,b)$. The leading $(m+1)\times(m+1)$ minor of $H$ is a polynomial in $g(0),\dots,g(2m)$. For $g(r)=\one[r=m]$ the leading block is the reversal permutation matrix, whose determinant is $\pm1$, so the polynomial is not identically zero and $\rank H=m+1$ for all $g$ outside a Lebesgue-null set.

(b) Here $H_{st}=\one[s+t\ge q]$, and row $s$ is the indicator of $\{t\ge c(s)\}$ with $c(s)=\max(q-s,0)$. Rows with $c(s)>b$ vanish. The nonzero rows are step vectors $\one[t\ge c]$, and distinct step vectors are linearly independent, since their matrix is triangular after ordering by $c$. Hence $\rank H$ is the number of distinct values of $c(s)$ over $\max(0,q-b)\le s\le a$. The function $c$ decreases strictly until it reaches $0$ at $s=q$ and is constant afterwards, so the number of distinct values is $\min(a,q)-\max(0,q-b)+1$. Checking the four cases $q\le a$ or $q>a$ and $q\le b$ or $q>b$, and using $a+b=N$, this equals $1+\min(q,N-q,a,b)$. For the dividends, $d_T=\sum_{U\subseteq T}(-1)^{|T|-|U|}\one[|U|\ge q]=\sum_{u=q}^{t}(-1)^{t-u}\binom tu$ with $t=|T|$. For $t<q$ the sum is empty. For $t\ge q\ge1$ the identity $\sum_{u=0}^{r}(-1)^u\binom tu=(-1)^r\binom{t-1}{r}$, which follows from $\binom tu=\binom{t-1}{u}+\binom{t-1}{u-1}$ by telescoping, applied with $r=q-1$ and $\sum_{u=0}^t(-1)^u\binom tu=0$, gives $d_T=(-1)^{t-q}\binom{t-1}{q-1}\ne0$.

(c) Here $H_{st}=\one[s+t=w]$, a partial permutation matrix whose rank is the number of $s\in[0,a]$ with $0\le w-s\le b$, that is, $\min(a,w)-\max(0,w-b)+1=1+\min(w,N-w,a,b)$ by the same case analysis as in (b).

(d) Here $H_{st}=\min(s,t)$. Row and column $0$ vanish. Let $m=\min(a,b)$. Since $\min(s,t)=\sum_{r=1}^{m}\one[r\le s]\one[r\le t]$ for $1\le s,t\le m$, the block $(\min(s,t))_{1\le s,t\le m}$ equals $UU^{\top}$ with $U$ the lower-triangular matrix of ones, whose determinant is one. Hence $\rank H=m$. For the dividends, let $T\cap L=A$ and $T\cap R=B$ with $|A|=\alpha$ and $|B|=\beta$. If $\alpha=0$ or $\beta=0$, every coalition inside $T$ contains no pair, so $v$ vanishes on subsets of $T$ and $d_T=0$. Otherwise write $\min(s,t)=\sum_{r\ge1}\one[s\ge r]\one[t\ge r]$. The M\"obius transform factorizes over $A$ and $B$, and the computation in (b) gives $\sum_{s}(-1)^{\alpha-s}\binom\alpha s\one[s\ge r]=(-1)^{\alpha-r}\binom{\alpha-1}{r-1}$. Hence
\[
d_T=(-1)^{\alpha+\beta}\sum_{r\ge1}\binom{\alpha-1}{r-1}\binom{\beta-1}{r-1}=(-1)^{\alpha+\beta}\binom{\alpha+\beta-2}{\alpha-1}\ne0
\]
by Vandermonde's identity.

\subsection{Theorem~\ref{thm:renyiall}}\label{app:renyi}

Let $\mu_k=\lambda_k^2$, so that $\mu_1\ge\mu_2\ge\cdots$, $\sum_k\mu_k=1$, and at most $n$ of the $\mu_k$ are nonzero. Let $t=\sum_{k\ge3}\mu_k$. The proof of Theorem~\ref{thm:renyi} gives $t\le\gamma$. The two largest of at most $n$ numbers that sum to one sum to at least $2/n$, so $t\le1-2/n$. Hence $t\le\gamma_n$. Define
\[
f(s)=2^{1-\alpha}(1-s)^\alpha+(n-2)^{1-\alpha}s^\alpha,\qquad s\in[0,1-2/n].
\]
For $n=2$ we have $t=0$ and $S^{(\alpha)}\le\ln2=\Phi_\alpha(\gamma;2)$ for every $\alpha$, so assume $n\ge3$.

(i) \emph{Case $\alpha\in(0,1)$.} The function $x\mapsto x^\alpha$ is concave. By Jensen's inequality $\mu_1^\alpha+\mu_2^\alpha\le2\bigl((1-t)/2\bigr)^\alpha$ and $\sum_{k=3}^n\mu_k^\alpha\le(n-2)\bigl(t/(n-2)\bigr)^\alpha$, so $\sum_k\mu_k^\alpha\le f(t)$. The derivative $f'(s)=\alpha\bigl[((n-2)/s)^{1-\alpha}-(2/(1-s))^{1-\alpha}\bigr]$ is nonnegative exactly when $(n-2)/s\ge2/(1-s)$, that is, when $s\le1-2/n$. So $f$ is nondecreasing on $[0,1-2/n]$, $f(t)\le f(\gamma_n)$, and $S^{(\alpha)}=\frac1{1-\alpha}\ln\sum_k\mu_k^\alpha\le\Phi_\alpha(\gamma;n)$. \emph{Case $\alpha\in(1,\infty)$.} Now $x^\alpha$ is convex, both Jensen inequalities reverse, and $\sum_k\mu_k^\alpha\ge f(t)$. The same computation shows that $f$ is nonincreasing on $[0,1-2/n]$, so $\sum_k\mu_k^\alpha\ge f(\gamma_n)$. Since $1/(1-\alpha)<0$, $S^{(\alpha)}\le\Phi_\alpha(\gamma;n)$. \emph{Case $\alpha=1$.} The function $x\mapsto-x\ln x$ is concave, so $-\mu_1\ln\mu_1-\mu_2\ln\mu_2\le-(1-t)\ln\frac{1-t}2$ and $-\sum_{k\ge3}\mu_k\ln\mu_k\le-t\ln\frac t{n-2}$. The sum of the right-hand sides is $g(t)=(1-t)\ln2+h(t)+t\ln(n-2)$, and $g'(s)=\ln\frac{(1-s)(n-2)}{2s}\ge0$ for $s\le1-2/n$, so $S^{(1)}\le g(\gamma_n)=\Phi_1(\gamma;n)$. \emph{Case $\alpha=\infty$.} $S^{(\infty)}=-\ln\mu_1$ and $\mu_1\ge(\mu_1+\mu_2)/2=(1-t)/2\ge(1-\gamma_n)/2$.

(ii) For $\alpha\in(1,\infty)$, removing the nonnegative second term inside the logarithm of~\eqref{eq:Phi} decreases the logarithm, and $1/(1-\alpha)<0$, so $\Phi_\alpha(\gamma;n)\le\frac1{1-\alpha}\ln\bigl(2^{1-\alpha}(1-\gamma_n)^\alpha\bigr)=\ln2-\frac{\alpha}{\alpha-1}\ln(1-\gamma_n)\le\ln2-\frac{\alpha}{\alpha-1}\ln(1-\gamma)$, because $\gamma_n\le\gamma$. For $\alpha=\infty$ the same holds with $\frac{\alpha}{\alpha-1}$ replaced by one. At $\alpha=2$ the bound is~\eqref{eq:thm4}.

(iii) Take the construction of Remark~\ref{rem:sharp}(b) with $n=2^{\min(|L|,|R|)}$, which needs $n$ orthonormal vectors on each side and $n\ge4$. Its normalized squared singular values are $(1-\gamma)/2$ twice and $\gamma/(n-2)$ with multiplicity $n-2$, and $\gamma_{\cut}=\gamma$. Since $\gamma\le1-2/n$ we have $\gamma_n=\gamma$, and $\sum_k\mu_k^\alpha=f(\gamma)$ for $\alpha\ne1$ and $-\sum_k\mu_k\ln\mu_k=g(\gamma)$ for $\alpha=1$, irrespective of the order of the singular values. Hence $S^{(\alpha)}_{\cut}=\Phi_\alpha(\gamma;n)$ for every $\alpha\in(0,\infty)$. For $\alpha=\infty$, $\gamma\le1-2/n$ gives $\gamma/(n-2)\le1/n\le(1-\gamma)/2$, so $\mu_1=(1-\gamma)/2$ and $S^{(\infty)}_{\cut}=\ln2-\ln(1-\gamma)=\Phi_\infty(\gamma;n)$.

\subsection{Proposition~\ref{prop:sii}}\label{app:sii}

(i) For $j\ge0$, $1/(j+1)=\int_0^1x^j\,dx$. Hence $\ISh(T)=\int_0^1\sum_{U\supseteq T}x^{|U|-|T|}d_U\,dx$. The integrand is the entry $T$ of $K_x^{\otimes N}\D$, because $\prod_i(K_x)_{t_iu_i}$ equals $x^{|U\setminus T|}$ if $T\subseteq U$ and zero otherwise. This gives $\mathcal I=\int_0^1K_x^{\otimes N}\D\,dx$. Every entry of the integrand is a polynomial in $x$ of degree at most $N$. The Gauss--Legendre rule on $[0,1]$ with $q=\lceil(N+1)/2\rceil$ nodes $x_1,\dots,x_q$ and positive weights $\omega_1,\dots,\omega_q$ integrates polynomials of degree at most $2q-1\ge N$ exactly, so $\mathcal I=\sum_{j=1}^q\omega_jK_{x_j}^{\otimes N}\D$. The unfolding of each term is $K_{x_j}^{\otimes|L|}D_{\cut}(K_{x_j}^{\otimes|R|})^{\top}$, which has rank $\chi_{\cut}$ because $\det K_x=1$. Subadditivity of rank gives $\rank I_{\cut}\le q\chi_{\cut}$. For the sharper bound write $K_x^{\otimes|L|}=\sum_{j=0}^{|L|}x^jA_j$, where $(A_j)_{T_L,U_L}=\one[T_L\subseteq U_L,\ |U_L\setminus T_L|=j]$. Then $I_{\cut}=\sum_{j=0}^{|L|}A_jD_{\cut}M_j$ with $M_j=\int_0^1x^j(K_x^{\otimes|R|})^{\top}dx$, so $\rank I_{\cut}\le(|L|+1)\chi_{\cut}$, and by symmetry $\rank I_{\cut}\le(|R|+1)\chi_{\cut}$. Since $\min(|L|,|R|)\le\lfloor N/2\rfloor$, this bound is never larger than $q\chi_{\cut}$, and item (ii) with $T=[N]$ attains it.

(ii) For $u_T$ the only nonzero dividend is $d_T=1$, so $\ISh(U)=\one[U\subseteq T]/(|T|-|U|+1)$. Write $U_L=U\cap L$ and $U_R=U\cap R$. The entry $(U_L,U_R)$ of $I_{\cut}$ is nonzero only if $U_L\subseteq T\cap L$ and $U_R\subseteq T\cap R$, and then it equals $1/(s+t+1)$ with $s=a-|U_L|$ and $t=b-|U_R|$. Hence $I_{\cut}=A_LHA_R^{\top}$ with $H_{st}=1/(s+t+1)$ for $0\le s\le a$, $0\le t\le b$, and with $(A_L)_{U_L,s}=\one[U_L\subseteq T\cap L,\ a-|U_L|=s]$ and $A_R$ defined in the same way. The columns of $A_L$ and $A_R$ have disjoint nonempty supports, so $\rank I_{\cut}=\rank H$. The leading $(m+1)\times(m+1)$ block of $H$, with $m=\min(a,b)$, is the Hilbert matrix of order $m+1$, which is nonsingular by Cauchy's determinant formula. Hence $\rank H=m+1$, while $\chi_{\cut}=1$ by Corollary~\ref{cor:exact}(d).

(iii) The dividend formula is inverted by $d_T=\sum_{U\supseteq T}B_{|U|-|T|}\ISh(U)$, where $B_0=1$, $B_1=-1/2$, $B_2=1/6,\dots$ are the Bernoulli numbers~\citep{grabisch2000}. The weight depends only on $|U\setminus T|=|U_L\setminus T_L|+|U_R\setminus T_R|$. Grouping the terms by $j=|U_L\setminus T_L|$ with the matrices $A_j$ of (i) gives $D_{\cut}=\sum_{j=0}^{|L|}A_jI_{\cut}M'_j$ with $(M'_j)_{U_R,T_R}=B_{j+|U_R\setminus T_R|}\one[T_R\subseteq U_R]$. Hence $\chi_{\cut}=\rank D_{\cut}\le(|L|+1)\rank I_{\cut}$, and by symmetry $\chi_{\cut}\le(|R|+1)\rank I_{\cut}$. For the equality case note that the map from dividends to $\ISh$ is invertible, so there is a game with $\ISh(U)=\one[U=[N]]$. Its dividends are $d_T=B_{N-|T|}=B_{s+t}$ with $s=|L|-|T_L|$ and $t=|R|-|T_R|$. As in (ii), $D_{\cut}=A'_LHA_R'^{\top}$ with the Hankel matrix $H_{st}=B_{s+t}$ for $0\le s\le|L|$, $0\le t\le|R|$, where $(A'_L)_{T_L,s}=\one[|L|-|T_L|=s]$ and $A'_R$ is defined in the same way, so $\chi_{\cut}=\rank H$, while $\rank I_{\cut}=1$. It remains to show that the leading block $H_m=(B_{s+t})_{0\le s,t\le m}$ with $m=\min(|L|,|R|)$ is nonsingular. Let $X$ be a real random variable with the logistic density $\frac{\pi}{2}\cosh^{-2}(\pi x)$, whose characteristic function is $\mathbb E\,e^{iuX}=(u/2)/\sinh(u/2)$, analytic for $|u|<2\pi$. Then $\mathbb E\,e^{z(iX-1/2)}=e^{-z/2}(z/2)/\sinh(z/2)=z/(e^z-1)$ for $|z|<2\pi$, so $B_n=\mathbb E\,(iX-\tfrac12)^n$ for every $n$, and $B_{s+t}=\mathbb E\,[(iX-\tfrac12)^s(iX-\tfrac12)^t]$. Expanding both powers gives $H_m=P\,\bigl(i^{k+l}\mu_{k+l}\bigr)_{0\le k,l\le m}P^{\top}$ with the moments $\mu_n=\mathbb E\,X^n$ and the unit lower triangular matrix $P_{sk}=\binom sk(-\tfrac12)^{s-k}$. Hence $\det H_m=(-1)^{m(m+1)/2}\det(\mu_{k+l})_{0\le k,l\le m}$. The moment matrix is positive definite, because $\sum_{k,l}c_kc_l\mu_{k+l}=\mathbb E\bigl(\sum_kc_kX^k\bigr)^2>0$ for $c\ne0$, as $X$ has a density. So $\rank H=m+1=(m+1)\rank I_{\cut}$. The script confirms $\det H_m\ne0$ for $m\le\BernHankelM$ in exact rational arithmetic, where it equals $(-1)^{m(m+1)/2}\prod_{k=1}^m(k!)^6/\bigl((2k)!\,(2k+1)!\bigr)$.

\end{document}